\documentclass[12pt]{article}
\usepackage[utf8]{inputenc}
\usepackage[margin=1in]{geometry}
\usepackage{setspace}
\usepackage{amsmath}
\usepackage{amssymb}
\usepackage{algorithm}
\usepackage{algpseudocode}
\allowdisplaybreaks
\usepackage{hyperref}
\makeatletter
\newcommand{\printfnsymbol}[1]{%
  \textsuperscript{\@fnsymbol{#1}}%
}
\makeatother
\usepackage{graphicx}
\usepackage{caption}
\usepackage{natbib}
\usepackage{subcaption}
\usepackage{color}
\usepackage{bbm}
\usepackage{float}

\usepackage{amsthm}
\usepackage{dsfont}
\usepackage{mathrsfs}
\usepackage{lmodern}
\usepackage{defs}
\usepackage{defs_general}
\usepackage{xcolor}

\usepackage{appendix}
\title{Independent and Decentralized Learning in Markov Potential Games\thanks{This version: December, 2024.}}
\author{Chinmay Maheshwari\printfnsymbol{2}, Manxi Wu, Druv Pai, and Shankar Sastry \thanks{
C. Maheshwari (chinmay\_maheshwari@berkeley.edu), D. Pai (druvpai@berkeley.edu), and S. Sastry (shankar\_sastry@berkeley.edu) are with the Department of Electrical Engineering and Computer Sciences at the University of California, Berkeley. M. Wu (manxiwu@cornell.edu) is with the Department of Civil and Environmental
Engineering, University of California at Berkeley.}}
\date{}

\begin{document}

\maketitle
\begin{abstract}
We study a multi-agent reinforcement learning dynamics, and analyze its asymptotic behavior in infinite-horizon discounted Markov potential games. We focus on the independent and decentralized setting, where players do not know the game parameters, and
cannot communicate or coordinate.
In each stage, players update their estimate of Q-function that evaluates their total contingent payoff based on the realized one-stage reward in an asynchronous manner. Then, players independently update their policies by incorporating an optimal one-stage deviation strategy based on the estimated Q-function. Inspired by the actor-critic algorithm in single-agent reinforcement learning, a key feature of our learning dynamics is that agents update their Q-function estimates at a faster timescale than the policies. Leveraging tools from two-timescale asynchronous stochastic approximation theory, we characterize the convergent set of learning dynamics.
\end{abstract}

\section{Introduction}
Markov games are a useful framework for modeling the strategic interactions among multiple self-interested players in a dynamic environment, where the utilities and state transitions are jointly determined by the actions of all players. {This framework has been adopted to study many} important applications that include
autonomous driving \cite{shalev2016safe}, adaptive traffic control \cite{prabuchandran2014multi,bazzan2009opportunities}, e-commerce \cite{kutschinski2003learning}, and AI training in real-time strategy games \cite{vinyals2019grandmaster,brown2018superhuman}. In these applications, players often need to act in an independent and decentralized manner, adapting to the information received through interactions in uncertain and dynamic environments. Coordination and communication may be absent, and players might not even be aware of the existence of others. In such an environment, a natural approach for each agent is to adopt a single-agent reinforcement learning algorithm, which uses only the information from the observed state, each player's individual action and bandit feedback about their own payoff in each stage. Such learning dynamics are fully decentralized and independent, meaning that each agent updates their own policy as if they are the sole decision-maker in the environment even though they are playing a Markov game. This leads to the following important question:

\medskip
\begin{quote}
\emph{What is the long run 
outcome of interactions among players who update their strategies in an independent and decentralized manner using RL algorithms?}
\end{quote}
\medskip

In this article, we study the above question in the context of Markov Potential Games (MPGs) (\cite{leonardos2021global,zhang2021gradient,song2021can,mao2021decentralized,ding2022independent,fox2022independent,zhangglobal,marden2012state,macua2018learning}). In MPGs, the change of utility of any player from unilaterally deviating their policy can be evaluated by the change of the potential function. 
{MPGs can be used to study Markov team games (also known as common interest games) \cite{arslan2016decentralized}, stochastic congestion games \cite{fox2022independent}, and dynamic demand-response in energy marketplaces \cite{narasimha2022multi}.} Previous studies of MPGs have mainly focused on analyzing convergence properties of gradient-based algorithms 
in both discounted infinite horizon settings \cite{zhang2021gradient,leonardos2021global,ding2022independent,fox2022independent} and finite horizon episodic settings \cite{mao2021decentralized,song2021can}.
However, the evaluation of gradient (or its estimate) of any player's value function 
requires players to either have access to a simulator/oracle of the value function or to coordinate and communicate their strategies and payoffs with each other \cite{leonardos2021global,fox2022independent, daskalakis2020independent,ding2022independent}. 
Such communication and coordination may be restricted in many real-world multi-agent systems due to communication constraints or privacy concerns \cite{kutschinski2003learning,nuti2011algorithmic,leslie2006generalised}.  

We focus on the learning dynamics, where each agent independently and decentralizely adopts an actor-critic algorithm \cite{konda1999actor} with asynchronous step sizes.  
In our setting, players do not know the existence of other players participating in the game, and do not have knowledge of state transition probabilities, their own payoff functions or the opponents' payoff functions. {Additionally, players do not have access to any information about the potential function or its existence.} Each player \emph{only} observes the realized state, and their own realized payoff in each stage. 
In particular, the multi-agent actor-critic learning dynamics considered in this article has the following key features: 
\begin{itemize}
    \item[(i)] The dynamics have \emph{two timescales}: each player updates the q-estimate of their contingent payoff (represented as the Q-function defined in Sec. \ref{sec:model}) at a faster timescale, and update their policies at a slower timescale.
    \item[(ii)] Players are \emph{self-interested} in that their updated policy incorporates an \emph{optimal one-stage deviation} that maximizes the expected contingent payoff derived from the current q-estimate. 
    \item[(iii)] Learning is \emph{asynchronous} and \emph{heterogeneous} among players. In every stage, only the q-estimate of the realized state-action pair, and the policy corresponding to the realized state are updated. The remaining elements in q-estimate and policy remain unchanged. Furthermore, the stepsizes of updating the element correspond to each state and action are heterogeneous across players, and are asynchronously adjusted according to the number of times a state and that player's action are realized. 
    \item[(iv)] {In each stage, players generate their actions by combining their updated policy with a uniform randomization (exploration) of all of their actions in order to learn the Q-function across all states. {The exploration  probability can be heterogeneous across players.}} 
\end{itemize}

{Independent and decentralized learning algorithms often do not converge \cite{tan1993multi,matignon2012independent,mazumdar2019policy}. We develop a new approach to characterize the convergent \textit{set} our learning dynamics.} Our approach involves non-trivial extensions of the analysis of single-agent reinforcement learning to MPG, and 
developing game-theoretic tools to utilize the equilibrium condition of the underlying Markov game. Specifically, we study the asymptotic behavior of discrete-time dynamics using an associated continuous-time dynamical system by exploiting the timescale separation between the updates of q-estimate and policy \cite{borkar2009stochastic, tsitsiklis1994asynchronous, perkins2013asynchronous}. In this dynamical system, the fast dynamics – update of the q-estimates – can be analyzed while viewing the policy updates (the slow dynamics) as static, and thus the q-estimate of each player converges to their Q-function {(Lemma \ref{lemma:d_fast}). 
Importantly, we show that, for every \(\epsilon>0\), the potential function always increases along the trajectory of the continuous-time dynamical system outside a set of \(\epsilon\)-stationary Nash equilibrium, given that the total exploration probability of all players is bounded by \(L\epsilon\), where \(L >0\) is a game dependent parameter (Lemma \ref{lemma:d_potential}). This key lemma allows us to characterize the convergent set of policies. Particularly, we show that the trajectories converge to the smallest super-level set of potential function that contains the \(\epsilon\)-Nash equilibrium set (Theorem \ref{theorem:independent}). Furthermore, under additional assumption on the potential function and Nash equilibrium set, we establish convergence to the set of \(\epsilon\)-stationary Nash equilibrium (Corollary \ref{cor: RefineConvergenceNew}). Finally, we validate the performance of our algorithm on a numerical example.}

\color{black}
\subsection{Additional related Works}
Apart from learning in MPGs, another line of work in multi-agent reinforcement learning focuses on the fully competitive setting of Markov zero-sum games \cite{daskalakis2020independent,sayin2020fictitious,sayin2021decentralized,alacaoglu2022natural,guo2021decentralized,perolat2015approximate}. Most articles in this line of works require players to either observe the opponents' rewards or actions \cite{alacaoglu2022natural, sayin2020fictitious,sayin2022fictitious}, or to coordinate in policy updates \cite{daskalakis2020independent,guo2021decentralized}.
The paper \cite{sayin2021decentralized}  proposed an independent and decentralized learning dynamics, and showed its convergence in Markov zero-sum games. The algorithm in \cite{sayin2021decentralized} also has timescale separation between policy update and value update, but is in reversed ordering compared to ours. That is, their dynamics update value function at a slower timescale, and update policies at a faster timescale, while our policy update is slower compared to the q-estimate update. Another difference is that our learning dynamics adopts a different stepsize adjustment procedure that allows players to update their step-sizes based on their own counters of states and actions heterogeneously. We emphasize that the convergence analysis in our paper is different from that in \cite{sayin2021decentralized} due to the differences in the two learning algorithms and the inherent difference between Markov zero-sum games and Markov potential games.\footnote{Our convergence proof builds on the existence of potential function and the convergence of fast q-learning. On the other hand, the proof of convergence in zero-sum Markov games depends on the Shapley iteration convergence.}

Two-timescale based algorithms have also been studied in other non-zero sum games \cite{borkar2002reinforcement,perolat2018actor,prasad2015two,arslan2016decentralized,leslie2006generalised}. {{}Specifically, \cite{leslie2006generalised} studied the two-timescale based algorithm for static games.}
The paper \cite{borkar2002reinforcement} proposed an actor-critic algorithm, and showed that certain weighted empirical distribution of realized actions converges to a \emph{generalized Nash equilibrium}. 
In \cite{arslan2016decentralized}, the authors presented an algorithm in the setting of \emph{acyclic Markov games}, which subsume Markov team games. However, the proposed algorithm require coordination amongst players.
The paper \cite{prasad2015two,perolat2018actor} proposed actor-critic algorithms with a \emph{fast} value function update -- based on temporal difference learning -- and a \emph{slow} policy update. In \cite{prasad2015two}, the gradient-based policy update requires the knowledge of opponents' rewards. The paper \cite{perolat2018actor} adopted a best-response based policy update that is similar to our learning dynamics, and proved its convergence in multistage games, which are a class of generalized normal form games with tree structures. Our algorithm is different from the one in \cite{perolat2018actor}. First,  we consider a uniform exploration policy which is different from \cite{perolat2018actor} where the authors consider perturbed or smoothed best response (similar to \cite{leslie2006generalised}).  Second, we consider updates with asynchronous step sizes that are adjusted based on counters of each state and each state-action pair while \cite{perolat2018actor} considers homogeneous step sizes.
The proof technique developed in \cite{perolat2018actor} exploits the special tree structure of multistage games, they cannot be applied in our setting. 
{{} Additionally, \cite{yongacoglu2023satisficing} proposes a {two-loop algorithm} where the policies are updated in the \emph{outer-loop} and the value functions are updated in the \emph{inner-loop}. Between any two updates of outer-loop, the algorithm makes multiple rounds of update of the inner-loop. The role of two loop algorithms here is to ensure stationarity in the learning environment by fixing the policy updates while learning the value function. This algorithm requires coordination among agents to decide the length of the inner loop.}

Finally, our results also advance the rich literature of learning in stateless potential game that includes continuous and discrete time best response dynamics \cite{monderer1996potential, swenson2018best}, fictitious play \cite{monderer1996fictitious, hofbauer2002global, marden2009joint}, replicator dynamics \cite{panageas2016average, hofbauer2003evolutionary}, no-regret learning \cite{heliou2017learning, krichene2014convergence}, and payoff-based learning \cite{cominetti2010payoff, leslie2006generalised}. {{}In particular, our learning dynamics share similar spirit with the payoff-based learning dynamics in stateless potential games \cite{cominetti2010payoff, leslie2006generalised}. In payoff-based learning, players update their payoff estimates based only on their own payoffs and adjust their mixed strategy using a best response.} In MPGs, the payoff estimates of different state-action pairs are updated asynchronously, and the best response becomes an optimal one-stage deviation policy. Therefore, our result is not a direct extension of stateless potential games as it involves using reinforcement learning tools to study long-run behavior.

\subsubsection*{Outline} Section \ref{sec:model} presents Markov potential games. 
We present our independent and decentralized learning dynamics, and the convergence results in Section \ref{sec:result}, {validate the performance of algorithm numerically in Section \ref{sec:numerics},} and conclude our work in Section \ref{sec: Conclusion}. We include the proofs of technical lemmas in the appendix. 

\section{Model}\label{sec:model}
Consider a \emph{Markov game} \(\game\) with discrete and infinite time horizon. We define the Markov game by tuple \(\game = \langle I, \stateSet, (\actSet_i)_{i\in\playerSet}, (\stagePayoff_i)_{i\in\playerSet}, \transition, \discount\rangle\), where $I$ is a finite set of players;
\(\stateSet\) is a finite set of states;
\(\actSet_i\) is a finite
   set of actions with generic member $a_i$ for each player \(i \in I\), and \(a=(a_i)_{i \in I}  \in \actSet = \times_{i\in \playerSet}
   \actSet_i\) is the action profile of
   all players;
\(\stagePayoff_i(s, a): S \times A \to \mathbb{R}\) is the one-stage
   payoff of player $i$ with state $s\in S$, and action profile $a \in \actSet$; We define \(u_{\max} = \max_{i,s,a}|u_i(s,a)|\); $\transition = (\transition(s'|s, a))_{s, s' \in S, a \in A}$ is the
   state transition matrix and $\transition(s'|s, a)$ is the probability that state changes from $s$ to $s'$ with action profile $a$; and
 \(\discount\in (0,1)\) is the discount factor.

   We denote \emph{a stationary Markov policy} \(\policy_i = (\policy_i(s, a_i))_{s \in S, a_i \in A_i} \in \Pi_i = \Delta(\actSet_i)^{|\stateSet|}\), where \(\policy_i(s, a_i)\) is the probability that player $i$ chooses action $a_i$ given state $s$. For each $i \in I$ and each $s \in S$, we denote $\pi_i(s)=(\pi_i(s, \ai))_{\ai \in \Ai}$. The joint policy profile is denoted as \(\policy = (\policy_i)_{i\in \playerSet} \in \Pi= \times_{i \in I} \Pi_i\). We also use the notation \(\policy_{-i} = (\policy_j)_{j \in I \setminus \{i\}} \in \Pi_{-i} = \times_{j \in I \setminus \{i\}} \Pi_j\) to refer to the joint policy of all players except for player $i$. For concise presentation, we will use the following notation throughout the paper: \[
\stagePayoff_i(s,a_i,\policy_{-i}) =\sum_{a_{-i}}\policy_{-i}(s,a_{-i})\stagePayoff_i(s,a_i,a_{-i}),\] 
\[P(s'|s,a_i,\policy_{-i}) = \sum_{a_{-i}}\policy_{-i}(s,a_{-i}) P(s'|s,a_i,a_{-i}),\] and \[
P(s'|s,\policy) = \sum_{a_i}\sum_{a_{-i}}\policy_i(s,a_i)\policy_{-i}(s,a_{-i})  P(s'|s,a_i,a_{-i}).\]

    \medskip 
     The game proceeds in discrete-time stages indexed by \(k=\{0,1,...\}\). At \(k=0\), the initial state \(s^0\) is sampled from a distribution \(\initDist\). At every time step \(k\), given the state $s^k$, each player's action \(a^k_i \in \actSet_i\) is sampled from the policy $\pi_i(s^k)$, and the joint action profile is $a^k=(a^k_i)_{i \in I}$. The state transitions to \(s^{k+1}\sim P(\cdot|s^k,a^k)\) based on the current state $s^k$ and action profile $a^k$. Given an initial state distribution $\mu$, and a stationary policy profile $\pi$, the expected total discounted payoff of each player \(i \in I\) is given by:  
    \begin{align}\label{eq:V}
        \vFunc_{i}(\initDist, \pi) =
        \avg\ls{\sum_{k=0}^{\infty}\discount^k\stagePayoff_i(s^k,a^k)},
    \end{align}
    where \(s^0\sim \initDist\), \(a^k\sim \policy(s^k)\), and \(s^k\sim \transition(\cdot|s^{k-1},a^{k-1})\). For the rest of the article, with slight abuse of notation, we use \(\vFunc_{i}(s, \pi)\) to denote the expected total utility of player $i$ when the initial state is a fixed state $s \in S$. Thus, we have
    \(
        \vFunc_i(\mu, \policy) = \sum_{s \in S} \mu(s) \vFunc_i(s, \policy).
    \)

     \begin{definition}[Markov potential games \cite{leonardos2021global}]\label{def: PotentialGame} 
    A Markov game \(\game\) is a Markov potential game (MPG) if there exists a state-dependent potential function \(\potential: S \times \Pi\ra \mathbb{R}\) such that for every \(s \in S, i \in \playerSet\), \(\policy_i,\policy_i'\in \Pi_i\), \(\policy_{-i}\in \Pi_{-i}\),
    \[\Phi(s, \policy_i',\policy_{-i}) -  \Phi(s, \policy_i,\policy_{-i})  = \vFunc_i(s, \policy_i',\policy_{-i}) -  \vFunc_i(s, \policy_i,\policy_{-i}).\] Moreover, given an initial state distribution $\mu\in \Delta(S)$, the potential function $\Phi(\mu, \pi):= \sum_{s \in S} \mu(s) \Phi(s, \pi)$ satisfies 
\[
       \Phi(\mu, \policy_i',\policy_{-i}) -  \Phi(\mu, \policy_i,\policy_{-i})= \vFunc_i(\mu, \policy_i',\policy_{-i}) -  \vFunc_i(\mu, \policy_i,\policy_{-i}), \]
       for every \(i \in \playerSet\), \(\policy_i,\policy_i'\in \Pi_i\), \(\policy_{-i}\in \Pi_{-i}\). 
    \end{definition}
That is, in a MPG, the change of a single deviator's value function can be characterized by the change of the value of the potential function.

We next present the definition of stationary Nash equilibrium, and $\epsilon$-stationary Nash equilibrium.
\medskip 
\begin{definition}[Stationary Nash equilibrium policy]
    A policy profile \(\policyNash\) is a stationary Nash equilibrium of $\game$ if for any \(i\in \playerSet\), any \(\policy_i\in\Pi_i\), and any \(\initDist\in\Delta(\stateSet)\),
    \(
        \vFunc_i(\mu, \policyNash_i,\policyNash_{-i}) \geq \vFunc_i(\mu, \policy_i,\policyNash_{-i}).
    \)
    \end{definition}
    \medskip 
    
    \begin{definition}[$\epsilon$-Stationary Nash equilibrium policy]
    For any $\epsilon \geq 0$, a policy profile \(\policyNash\) is an $\epsilon$-stationary Nash equilibrium of $\game$ if for any \(i\in \playerSet\), any \(\policy_i\in\Pi_i\), and any \(\initDist\in\Delta(\stateSet)\),
    \(
        \vFunc_i(\mu, \policyNash_i,\policyNash_{-i}) \geq \vFunc_i(\mu, \policy_i,\policyNash_{-i}) -\epsilon.
    \)
    {For any \(\epsilon\geq 0,\) we define the set of all \(\epsilon\)-stationary Nash equilibrium as \(\textsf{NE}(\epsilon)\).}
    Any \(\epsilon\)-stationary Nash equilibrium with \(\epsilon=0\) is a Nash equilibrium.
    \end{definition}

Both stationary Nash equilibrium and $\epsilon$-stationary Nash equilibrium exist in Markov games with finite states and actions \cite{fudenberg1991game}. In a MPG, if there exists a policy \(\pi^\ast\) such that $\pi^\ast = \arg\max_{\pi \in \Pi} \Phi(s, \pi)$ for every \(s\in S\), then \(\pi^\ast\) is a stationary Nash equilibrium policy of the MPG. {However, computing the Nash equilibrium as the maximizer of $\Phi(s,\cdot)$ is impossible in our setting since the players do not have the knowledge of the potential function $\Phi(s, \cdot)$ or the oracle access to its value. Moreover, even in settings where the potential function is known, e.g. common interest games, computing its maximizer is challenging due to the fact that $\Phi(s, \pi)$ is non-linear and non-concave in $\pi$. }

\section{Independent and Decentralized Learning Dynamics
}\label{sec:result}
In this section, we present the learning dynamics and  {characterize its long-run behavior}. First, we highlight the information available to every player in the learning process.

{\textbf{Information available to players.}
We assume that each player $i \in I$ knows their own action set $A_i$ and the state set \(S\). Players do \emph{not} know the state transition probability matrix $P$, their own or others' payoff functions $(u_i)_{i \in I}$, and they do not know the initial state distribution \(\mu\in \Delta(S)\). Players do not know the existence of other players or the underlying potential function of the game. In each stage $k=0,1, 2, \dots$ of the learning algorithm, players observe the realized state $s^k$ that they use to compute the action \(a^k\) and in turn obtain the reward $r_i^k = u_i(s^k, a^k)$. We want to emphasize that the players only receive the bandit feedback of reward function, i.e. in any stage they do not receive the reward corresponding to the action they did not choose. Additionally, players do not observe the actions or the rewards of their opponents. Moreover, players do not know the parameters used by other players in the learning dynamics (to be presented shortly).
}

Given any policy $\pi \in \Pi$, and any initial state $s \in S$, we define the following \emph{Q-function} for each player $i \in I$ and action $a_i \in A_i$: 
    \begin{align}\label{eq:Q}
        Q_i(s, a_i; \pi)&= \stagePayoff_i(s,a_i, \pi_{-i}) + \delta \sum_{s'\in \stateSet} \transition(s'|s,a_i, \pi_{-i}) \vFunc_i(s', \pi). 
    \end{align}
    In \eqref{eq:Q}, player $i$'s expected utility in the 
    first stage
    with state $s$ is derived from playing action $a_i$ and her opponents choosing policy $\pi_{-i}$. The expected total utility starting from
    stage 2
    is derived from all players following policy $\pi$. Therefore, the Q-function $Q_i(s, a_i; \pi)$ represents player $i$'s expected discounted utility when the game starts in state $s$, and player $i$ deviates for \emph{one-stage} (namely, the first stage) from her policy to play $a_i$. With slight abuse of notation, we define \(Q_i(s;\pi) = (Q_i(s,a_i;\pi))_{a_i\in A_i}\in \R^{|A_i|}\) for every \(i\in I, s\in S, \pi\in \Pi\). Furthermore, we define \emph{optimal one-stage deviation} from policy \(\pi\in \Pi\) in state \(s\in S\) as 
  \begin{align}\label{eq: BestResponseExact}
    {\mathrm{br}}_i(s;\policy) &= \underset{\hat{\pi}_i\in \Delta(A_i)}{\arg\max}
    ~ \hat{\pi}_i^\top Q_i(s;\pi).
    \end{align}
    One can obtain equivalent characterization of Nash equilibrium in terms of optimal one-stage deviation. Specifically, a policy is a Nash equilibrium if and only if it is fixed point of optimal one-stage deviation (Lemma \ref{lem:FixedPoint}).
\subsection{Learning Dynamics}
The proposed learning happens in iterates, denoted by \(t\). In every iterate $t$, each player $i \in I$ updates four components \(\nk, \nkak^t, \localQTilde_i^t, \piik\). In particular, $\nk=(\nk(s))_{s \in S}$ is the vector of state counters, where $\nk(s)$ is the number of times state $s$ is realized before iterate $t$. For each player $i \in I$, $\nkak^t= (\nkak^t(s, \ai))_{s \in S, \ai \in \Ai}$ is the counter of state-action tuple, where $\nkak^t(s, \ai)$ is the number of times that player $i$ has played action $\ai$ in state $s$ before iterate $t$. 

Additionally, $\localQTilde_i^t= (\localQTilde_i^t(s, \ai))_{s \in S, \ai \in \Ai}$ is player $i$'s estimate of her Q-function, and $\pi_i^t= (\pi^t_i(s, \ai))_{s \in S, \ai \in \Ai}$ is player $i$'s policy in iterate $t$.  
Given the state-(local) action tuple of any player \(i\), $(s^{t-1}, a^{t-1}_i)$, the estimate $\localQTilde_i^t(s^{t-1}, a^{t-1}_i)$ is updated in \eqref{eq:d_q} as a linear combination of the estimate $\localQTilde_i^{t-1}(s^{t-1}, a^{t-1}_i)$ in the previous stage, and a new estimate that is comprised of the realized one-stage payoff $\reward_i^{t-1}$ and the estimated total discounted payoff from the next iterate. 

The policy $\pi_i^{t}(s^{t-1})$ updated in \eqref{eq:d_pi} is a linear combination of the policy $\pi_i^{t-1}(s^{t-1})$ in the previous stage, and player $i$'s optimal one-stage deviation. Particularly, the optimal one-stage deviation is computed using the updated q-estimate $\localQTilde_i^t$, instead of the actual ${Q}_i$, which is unknown to the player. Each player's action $a_i^t$ is sampled from their policy \(\pi_i^{t}\) with probability $(1-\theta_i)$ and from a uniform distribution over their action set $A_i$ with probability $\theta_i$, where \(\theta_i\in (0,1)\) is the exploration parameter that can be heterogeneous among players (refer to \eqref{eq: ActionSampling}). With slight abuse of notation, we define \(\theta = (\theta_i)_{i\in \mathcal{I}}\in (0,1)^{|\playerSet|}.\)

Note that the updates of $\localQTilde_i^t$ (resp. $\pi_i^{t}$), in each iterate, only change the element that corresponds to the realized state and action $(s^{t-1}, a^{t-1})$ (resp. state $s^{t-1}$), and the remaining elements stay unchanged. 
Furthermore, the update speed of $\localQTilde_i^t(s^{t-1}, \ai^{t-1})$ (resp. $\policy_i^t(s^{t-1})$) is governed by the step size sequence $\alpha_i(n)$ (resp. $\beta_i(n)$) corresponds to the state-action counter $\tilde{n}= \nkak^t(s^{t-1}, \ai^{t-1})$ (resp. state counter $n=n^t(s^{t-1})$). Therefore, the update is \emph{asynchronous} in that the stepsizes are different across the elements associated with different states and actions, {and stepsizes are different for different players.}

\begin{algorithm}[htp]
\textbf{Initialization:} $n^0(s)=0, \forall s \in\stateSet$; \(\tilde{n}_i^0(s, \ai)=0, \localQTilde_{i}^0(s, a_i)=0\), set arbitrary \(\pi_i^0(s, \ai), ~\forall (i, \ai, s)\), and \(\explore_i\in (0,1)\), \(\forall \ i\). 

In iterate 0, each player observes \(s^0\in S\), choose their action $a_i^0 \sim \pi_i^0(s^0)$, and observe \(r_i^0=u_i(s^0,a^0)\). 

\textbf{In every iterate $t=1,2,...$,} each player observes $s^t$, and independently updates \(\{n^t,\nkak^t,\localQTilde_i^t,\policy_i^t\}\). \;

\smallskip
\textit{Update $\nk,\nkak^t$}: 
\begin{align*}
    &\nk(s^{t-1}) = n^{t-1}(s^{t-1})+1, \\ 
    &\nkak^{t}(s^{t-1},a^{t-1}_i) = \nkak^{t-1}(s^{t-1},a^{t-1}_i)+1, 
\end{align*}
Furthermore, for \(s\neq s^{t-1},a_i\neq a_i^{t-1}\), 
{\begin{align*}
     \nkak^{t}(s,a_i)=\nkak^{t-1}(s,a_i), \quad \nk(s) = n^{t-1}(s) .
\end{align*}}

\textit{Update $\localQTilde_i^t$}:
{\small\begin{subequations}\label{eq:d_q}
\begin{align}
   & \localQTilde_{i}^{t}(s^{t-1},a^{t-1}_i) = \localQTilde_{i}^{t-1}(s^{t-1},a_i^{t-1})+ \stepFast_i(\nkak^{t}(s^{t-1},a_i^{t-1})) \notag\\
    &\quad  \cdot\bigg( \reward_i^{t-1}+\discount 
   \pi_i^{t-1}(s^{t})^\top  \localQTilde_i^{t-1}(s^t)
    - \localQTilde_i^{t-1}(s^{t-1},a_i^{t-1})\bigg), \\ 
    &\quad \localQTilde_i^{t}(s,a_i) = \localQTilde_i^{t-1}(s, a_i), \quad \forall \ (s,a_i)\neq (s^{t-1},a^{t-1}_i),
\end{align}
\end{subequations}}where \(r_i^{t-1}= u_i(s^{t-1},a^{t-1})\). 
\smallskip

{\textit{Update \(\policy_i^{t}\):} 
{\begin{subequations}\label{eq:d_pi}
{\begin{align}
    &\policy_i^{t}(s^{t-1}) =\policy_i^{t-1}(s^{t-1})+ \beta_i(\nk(s^{t-1}))\cdot\notag\\
    &\lr{\widehat{\br}_i^{t-1}(s^{t-1}) - \policy_i^{t-1}(s^{t-1}) }, \\ 
    &\policy_i^t(s) = \policy_i^{t-1}(s), \quad \forall \ s\neq s^{t-1},
\end{align}}
where \(\widehat{\br}_i^{t}(s) \in \arg\max_{\pi_i\in\Delta(A_i)}\pi_i^\top q_i^{t}(s)\) for every \(s\in S\). 
\end{subequations}}}

\smallskip 

{At the end of iterate $t$, each player chooses their action 
\begin{align}\label{eq: ActionSampling}
    a_i^t \sim (1-\explore_i)\pi_i^t(s^t)+ \explore_i \cdot (1/|A_i|)\mathbbm{1}_{A_i},
\end{align}where \(\mathbbm{1}_{A_i} \in \mathbb{R}^{|A_i|}\) is a vector with all entries 1. \\
Each player observes their own reward $r_i^t=u_i(s^t,a^t)$. }
\caption{Independent and decentralized learning dynamics}
\label{alg:independent_decentralized}
\end{algorithm}

\subsection{Convergence Analysis}
Next, we introduce the assumptions that are needed for obtaining {the convergent set of the learning dynamics}.

{
\begin{assumption}\label{as:basic}
The initial state distribution $\mu(s)>0$ for all $s \in S$. Additionally, \(\min_{s,s'\in S, a\in A} P(s'|s,a) > 0\). 
\end{assumption}
}
\smallskip
Assumption \ref{as:basic} ensures that every state is visited infinitely often  so that agents can learn the Q-function associated with each state. {We note that similar assumptions on the probability transition are commonly made in multi-agent reinforcement learning literature. For example, the paper \cite{leslie2020best} also assumed that \(\min_{s,s'\in S, a\in A}P(s'|s,a) > 0\) and \cite{sayin2021decentralized} assumed that for any pair of states \((s,s')\in S\times S\) and any infinite sequence of joint actions, the state \(s'\) is reachable from \(s\) in finite steps.}

Next, we make the following assumption on the stepsizes $\{\{\alpha_i(n)\}_{i\in\playerSet}\}_{n=0}^{\infty}$, $\{\{\beta_i(n)\}_{i\in\playerSet}\}_{n=0}^{\infty}$:

\smallskip 
\begin{assumption}\label{as:stepsize}
The step sizes \(\{\stepFast_i(n)\in(0,1)\}_{n=0,i\in \playerSet}^{\infty}\) and \(\{\stepSlow_i(n)\in(0,1)\}_{n=0, i\in \playerSet}^{\infty}\) satisfy 
\begin{itemize}
    \item[(i)] For all \(i\in \playerSet\), \(\sum_{n=0}^{\infty}\stepFast_i(n)=\infty, \sum_{n=0}^{\infty}\stepSlow_i(n)=\infty\), \(\lim_{n\ra\infty}\stepFast_i(n)=\lim_{n\ra\infty}\stepSlow_i(n)=0\);
    \item[(ii)] For every \(i\in \playerSet\), there exist some \(q,q'\geq 2\), \(\sum_{n=0}^{\infty}\stepFast_i(n)^{1+q/2}<\infty\) and \(\sum_{n=0}^{\infty}\stepSlow_i(n)^{1+q'/2}<\infty\);
    \item[(iii)] For every \(i\in \playerSet, x\in (0,1)\), \(\sup_n \stepFast_i([xn])/\stepFast_i(n) < A_x < \infty\), \(\sup_n \stepSlow_i([xn])/\stepSlow_i(n) < A_x < \infty\), {where \([xn]\) denotes the largest integer less than or equal to \(xn\)}. Additionally, \(\{\stepFast_i(n)\},\{\stepSlow_i(n)\}\) are non-increasing in \(n\);
    \item[(iv)] For every \(i,j\in \playerSet\),  \(\lim_{n\ra\infty}\stepSlow_i(n)/\stepFast_j(n) = 0\);
    {\item[(v)] For every \(i,j\in \playerSet\), there exists  \(0<\xi_{ij}^\alpha<  \zeta_{ij}^\alpha < \infty\), and \(0<\xi_{ij}^\beta< \zeta_{ij}^\beta < \infty\) such that  \(\frac{\stepFast_i(n)}{\stepFast_j(n)}\in [\xi_{ij}^\alpha, \zeta_{ij}^\alpha],\) and \(\frac{\stepSlow_i(n)}{\stepSlow_j(n)} \in [\xi_{ij}^\beta, \zeta_{ij}^\beta]\) 
    for all $n$.
    }
\end{itemize}
\end{assumption}
\smallskip

Assumption \ref{as:stepsize}(i) { ensures that the asymptotic properties of our learning dynamics can be studied by using a continuous-time dynamical system} \cite{borkar2009stochastic,benaim1999dynamics,tsitsiklis1994asynchronous}. {Assumption \ref{as:stepsize}(ii) 
ensures that the average cumulative impact of noise terms on the asymptotic behavior of learning dynamics diminishes. 
} Assumption \ref{as:stepsize}(iii) {
is a technical condition required for the asynchronous update in the learning dynamics to 
ensure that the value functions and policies associated with different state-action pairs are updated at the same timescale \cite{perkins2013asynchronous}.
}
Assumption \ref{as:stepsize}(iv) implies that our learning dynamics have two timescales: the update of $\{\localQTilde_i^t\}_{t=0}^{\infty}$ is asymptotically faster than the update of $\{\policy_i^t\}_{t=0}^{\infty}$. 
Assumption \ref{as:stepsize}(v) suggests that players can employ varying step sizes, provided that the ratio between their step sizes remains bounded between zero and a finite number for all steps.
One example of stepsizes that satisfies Assumption \ref{as:stepsize} is \(\alpha_i(n) =z_in^{-c_1}\) and \(\beta_i(n)= y_in^{-c_2}\) where \(0< c_1\leq c_2\leq 1\) and \(y_i,z_i\) {can be any} player specific positive scalars.

{Next, we state the main result of this paper that characterizes the convergent set of policy updates of Algorithm \ref{alg:independent_decentralized} as a super-level set of potential function \(\Phi\) (Theorem \ref{theorem:independent}). Furthermore, by imposing additional assumption on the potential function and the set of Nash equilibrium, we characterize the convergent set as a set of approximate Nash equilibrium (Corollary \ref{cor: RefineConvergenceNew}).

\begin{theorem}\label{theorem:independent}
Under Assumptions \ref{as:basic} and \ref{as:stepsize}, for every \(\epsilon > 0\), 
 the policy sequence $\{\pi^t\}_{t=0}^{\infty}$ induced by Algorithm \ref{alg:independent_decentralized} converges to the set  \begin{align}\label{eq: SetNotation}
 \Pi^\ast_\epsilon \defas \{\pi\in \Pi: \Phi(\mu,\pi)\geq \min_{\pi'\in \textsf{NE}(\epsilon)}\Phi(\mu,\pi')\}
 \end{align} with probability 1 given that 
\begin{align}\label{eq:theta_constraint}
     \sum_{i\in \playerSet}\theta_i < \epsilon \cdot \frac{ \eta (1-\discount)^2}{4u_{\max} |\playerSet|(\eta+(D/(1-\discount)))} =: L\epsilon,
 \end{align}
 where 
 {\small \begin{align*}
     &D = \frac{1}{1-\discount} \max_{i,\pi_{-i},s}\Big|\frac{d^{\pi_i^\dagger, \pi_{-i}}_{\mu}(s)}{\mu(s)}\Big|, \BR_i \in \arg\max_{\pi_i\in \Pi_i}{}{V}_i(\mu,\policy_i,\pi_{-i}), \\ 
     &\eta = \frac{\zeta}{A_{\zeta}}, \quad \zeta = \min_{s,s'\in S, a\in A}P(s'|s,a),
 \end{align*}}and $A_{\zeta}$ is defined as in Assumption \ref{as:stepsize} (iii). Moreover, for any \(\epsilon,\epsilon'\) such that \(0\leq \epsilon\leq \epsilon'\), \(\Pi^\ast_{\epsilon}\subseteq \Pi^\ast_{\epsilon'}\). Additionally, for every \(\epsilon\geq 0\), \(\textsf{NE}(\epsilon)\subseteq \Pi^\ast_{\epsilon}\).
\end{theorem}

Theorem \ref{theorem:independent}  guarantees that for any $\epsilon>0$, for sufficiently small exploration rate that satisfies \eqref{eq:theta_constraint}, the sequence of policies induced by the learning dynamics asymptotically converges to the smallest  super-level set of potential function that contains the set of \(\epsilon\)-stationary Nash equilibrium (i.e. \(\textsf{NE}(\epsilon)\)).   
{
Moreover, Theorem \ref{theorem:independent} states that for the policy sequence induced by Algorithm \ref{alg:independent_decentralized} to converge to  the set of approximate Nash equilibria with smaller approximation gap, the sum of the exploration probabilities of all players should be smaller.   
}

The convergence result of Theorem \ref{theorem:independent} can be refined to ensure convergence to an approximate Nash equilibrium set under additional Assumption \ref{assm: SublevelSetNew}.

{\begin{assumption}\label{assm: SublevelSetNew}
    For every \(\epsilon > 0\), there exists \(h_{\epsilon}\in \mathbb{R}_+\) such that \( \Pi^\ast_{\epsilon}\subseteq \textsf{NE}(\epsilon+h_{\epsilon}),\) \(h_{\epsilon}\) is continuous and non-decreasing in \(\epsilon\), and \( \lim_{\epsilon\downarrow 0}h_{\epsilon} = 0\).
\end{assumption}}

{{}\begin{corollary}\label{cor: RefineConvergenceNew}
    Suppose that Assumptions \ref{as:basic}, \ref{as:stepsize} and \ref{assm: SublevelSetNew} hold. 
 For every \(\tilde{\epsilon}, \tilde{\epsilon}'\) such that \(0 < \tilde{\epsilon} <   \tilde{\epsilon}',\) there exist positive scalars \(0<\epsilon< \epsilon'\) such that \({\epsilon}+h_{{\epsilon}} = \tilde{\epsilon}\) and \({\epsilon'}+h_{{\epsilon'}} =  \tilde{\epsilon}'\), and the sequence of policies $\{\pi^t\}_{t=0}^{\infty}$ induced by Algorithm \ref{alg:independent_decentralized} converges to the set \(\textsf{NE}(\tilde{\epsilon})\) (resp. \(\textsf{NE}(\tilde{\epsilon}')\))
 with probability 1, if \(\sum_{i\in I}\theta_i < L\epsilon\) (resp. \(\sum_{i\in I}\theta_i < L\epsilon'\)).
\end{corollary}
}
}

{{} Corollary \ref{cor: RefineConvergenceNew} states that for the policy sequence induced by Algorithm \ref{alg:independent_decentralized} to converge to the set of approximate Nash equilibria with a smaller approximation gap, the sum of the exploration probabilities of all players must be smaller. We have included the proof of Corollary \ref{cor: RefineConvergenceNew} in Appendix \ref{ssec: Corollaryproof}.}

{{} Now we prove Theorem \ref{theorem:independent}. First, we introduce some useful notations. For any \(i\in I, s\in S\), we define \(\pi_i^\circ(s)= (1/|A_{i}|)\cdot \mathbbm{1}_{A_{i}}\) to be a uniformly random policy. Additionally, for any \(\pi\in \Pi\) and any \(\theta = (\theta_i)_{i \in I} \in (0,1)^{|I|}\), we define \(\pi^{(\theta)}\in \Pi\) such that for every \(s\in S, i\in I\), \(\pi_{i}^{(\theta)}(s):= (1-\theta_i)\pi_{i}(s)+\theta_i\pi_i^\circ(s)\) to be a perturbed version of policy \(\pi\) given exploration probability vector \(\theta\).}

{To prove Theorem \ref{theorem:independent}, we apply the two-timescale asynchronous stochastic approximation theory \cite{perkins2013asynchronous}, where we first ensure the convergence of the (``fast'') q-estimate updates, \(\{\localQTilde_i^{t}\}_{t=0}^{\infty}\), in Lemma \ref{lemma:d_fast}. Next, in Lemma \ref{lemma:d_potential}-\ref{lemma: Policy_slow_continuous}, we study {{}the convergent set} of the (``slow'') policy updates given the convergent values of q-estimates.}

\medskip 
{{}\begin{lemma}\label{lemma:d_fast}
Under Assumptions \ref{as:basic} and \ref{as:stepsize}, for any $s \in S$ and $i \in I$, $\lim_{t\to \infty} \|\localQTilde_{i}^{t}(s) - {Q}_i(s; (\policy_i^t, \policy^{t,(\theta)}_{-i}))\|_\infty=0$ with probability 1.
\end{lemma}}
\medskip 

The proof of Lemma \ref{lemma:d_fast} is based on two steps. First, we show that under Assumption \ref{as:basic} and \ref{as:stepsize}, 
our learning dynamics satisfies the set of conditions introduced in \cite{perkins2013asynchronous} (restated in Appendix \ref{sec: PerkinsSA}) so that convergence of q-estimates can be analyzed by the associated continuous-time dynamical system where the policy drifts are treated as asymptotically negligible errors. Second, we argue the global convergence of the continuous-time dynamical system using the contraction property of the Bellman operator associated with q-estimate update. The complete proof of Lemma \ref{lemma:d_fast} is deferred to Appendix \ref{ssec: d_fast_lemma}. 

Next, we analyze the policy updates with respect to the convergent values of the q-estimates provided by the fast dynamics as in Lemma \ref{lemma:d_fast}. {{}Particularly, the policy $\pi_i^t(s^{t-1})$ in \eqref{eq:d_pi} becomes a linear combination of $\pi_i^{t-1}(s^{t-1})$, and the optimal one-stage deviation $\br_i(s^{t-1}; (\policy_i^{t-1},\policy_{-i}^{t-1,(\theta)}))$ based on the actual Q-function as in \eqref{eq: BestResponseExact}.} Under Assumption \ref{as:basic}, the asymptotic behavior of \(\{\policy^t\}_{t=0}^{\infty}\) can be analyzed using the following continuous-time differential inclusion, where $\tau\in [0, \infty)$ is a continuous-time index,
{{}\begin{align}\label{eq: diff_inclusion}
    &\frac{d}{d\tau}{\varpi}_i^\tau(s) \in \gamma_{i}(s) \lr{ \br_i(s;(\varpi_i^{\tau},\varpi_{-i}^{\tau,(\theta)})) -\varpi_i^\tau(s)}, 
\end{align}}and $\gamma_{i}(s)\in [\eta,1]$, for every \(s\in S, i\in I,\) captures \footnote{{} From two-timescale asynchronous stochastic approximation theory \cite{perkins2013asynchronous} (also reviewed in Appendix \ref{sec: PerkinsSA}), the value of \(\gamma_i(s) \geq \kappa/ A_{\kappa}\), where \(\kappa\) is a lower bound on the stationary distribution of the Markov chain over the state space induced by any policy. Assumption \ref{as:basic} ensures that the probability of every state in this stationary distribution, under any policy, is greater than \(\zeta\).
} the asynchronous update of policies in different states (cf. \eqref{eq:d_pi}), and  $\eta = \zeta/A_\zeta>0$ \cite{perkins2013asynchronous}. Since \(\br_i(\cdot)\) is a non-empty closed, convex and compact set, there exists an absolutely continuous solution of \eqref{eq: diff_inclusion}, \(\varpi^{\tau}_i\) for every \(i\in I\) \cite{benaim2005stochastic}.   
{{}Consequently, for every \(i\in I\) and \(s\in S\), there exists \(\widetilde{\mathrm{br}}^{\tau,(\theta)}_i(s) \in {\mathrm{br}}_i(s;(\varpi_i^{\tau},\varpi_{-i}^{\tau,(\theta)}))\) such that
\begin{align}\label{eq: policy_diff}
   \frac{d}{d\tau}{\varpi}_i^{\tau}(s) =  \gamma_{i}(s) \lr{
\widetilde{\mathrm{br}}^{\tau,(\theta)}_i(s)
     -\varpi_i^{\tau}(s)}.
 \end{align}  
}

To establish the convergence of \eqref{eq: policy_diff}, we define a Lyapunov function $\phi: [0, \infty) \to \mathbb{R}$ as follows:
\begin{align}\label{eq:phi_t}
    \phi(\tau) = \max_{\varpi\in \Pi}\sum_{s\in S}\mu(s){\Phi}(s,\varpi)-\sum_{s \in S} \mu(s) {\Phi}(s, \varpi^\tau),
\end{align}
which is the difference of the potential function at its maximizer with that of its value at \(\varpi^\tau\). 
We show that \(\phi(\tau)\) is strictly decreasing (alternatively, the potential function is strictly increasing) as long as \(\varpi^{\tau}\) is outside the set \(\textsf{NE}(\epsilon)\).

 \medskip 
\begin{lemma}\label{lemma:d_potential}
{{} Suppose that Assumptions \ref{as:basic} holds and \(\theta\) satisfies \eqref{eq:theta_constraint}. For every \(\tau\geq 0\) and  \(\epsilon > 0\), if \(\varpi^\tau \not\in \textsf{NE}(\epsilon)\) then \(d\phi(\tau)/d\tau < 0\). 
}
\end{lemma}
\medskip

Unlike static potential games, the potential function is non-concave in each player's policy in Markov potential games. Therefore, we need a new approach to demonstrate that $\phi(\tau)$ decreases outside a neighborhood of approximate Nash equilibrium. 
To prove Lemma \ref{lemma:d_potential}, we need the following technical lemma that extends the single-agent reinforcement learning theory to multi-agent games. 
\smallskip
\begin{lemma}\label{lem: TechnicalMain} 
\noindent (a) \emph{Gradient of value function}: For any \(\mu\in\Delta(\stateSet), s\in\stateSet,\policy\in\Pi,i\in\playerSet,a_i\in\Ai\),
\begin{align*}
    \frac{\partial {V}_i(\mu,\policy)}{\partial\policy_i(s,a_i)}&=\frac{1}{1-\discount}d^{\pi}_{\mu}(s){Q}_i({s},a_i;\policy),
\end{align*}
where 
\begin{align}\label{eq: d_pi_eq}
   d^{\pi}_{\mu}(s) \defas  (1-\discount)\sum_{s^0 \in S} \mu(s^0)\sum_{k=0}^{\infty}\discount^k\Pr(s^k=s|s^0). 
\end{align}

\smallskip 
\noindent (b) \emph{Multi-agent performance difference lemma}: For any policy \(\policy=(\policy_i,\policy_{-i}), \policy'=(\policy_i',\policy_{-i})\in \Pi\) and any \(\mu\in\Delta(\stateSet)\), 
{\small
\begin{align}\label{eq:performance_differenceMain}
&{\vFunc}_i(\mu,\policy)-{\vFunc}_i(\mu,\policy') = \frac{1}{1-\discount}\sum_{s'}d^{\pi}_{\mu}(s') \advFunc_i(s',\policy_i;\policy'),
\end{align}}where $\advFunc_i(s,a_i;\policy)$ is the \emph{advantage function} given by 
\begin{align}\label{eq:advantageMain}
    \advFunc_i(s,a_i;\policy) \defas {Q}_i(s,a_i;\policy)-{V}_i(s,\policy), 
\end{align}
for every \( i \in I, ~ s \in S, ~ \ai \in \Ai, \pi \in \Pi\).
\smallskip

{{}
\smallskip
\noindent (c)
\textit{Sensitivity of value function:} For any \(i\in I, \pi_i\in \Pi_i, \pi_{-i} \in\Pi_{-i}\), 
    \begin{align*}
        \max_{s\in S }|V_i(s,\pi_i,\pi_{-i}) - V_i(s,\pi_i,\pi_{-i}^{(\theta)})| \leq \frac{2\sum_{k\in \playerSet\backslash\{i\}}\theta_k}{(1-\discount)^2}u_{\max}.
    \end{align*}

\smallskip
\noindent (d) \textit{Sensitivity of Q-function:}    For any \(i\in I, \pi_i\in \Pi_i, \pi_{-i}\in \Pi_{-i}\), it holds that 
    \begin{align*}
\max_{s,a_i}|Q_i(s,a_i;\pi) -Q_i(s,a_i;\pi_i,\pi_{-i}^{(\theta)})  | \leq \frac{2\sum_{k\in \playerSet\backslash\{i\}}\theta_k}{(1-\discount)^2}u_{\max}.
    \end{align*}}
\end{lemma}
The proof of Lemma \ref{lem: TechnicalMain} is included in Appendix \ref{ssec: LemmaTechnicalMain}. We are now ready to prove Lemma \ref{lemma:d_potential} based on Lemma \ref{lem: TechnicalMain}. 

\medskip

\noindent{\emph{Proof of Lemma \ref{lemma:d_potential}}:}
We compute the derivative of $\phi(\tau)$ with respect to $\tau$, where $\phi(\tau)$ is given by \eqref{eq:phi_t}.  
\begin{align}
    &\frac{d}{d\tau}\phi(\tau) = -\sum_{i, s}\frac{\partial {}{\potential}(\mu,\varpi^{\tau}) }{\partial \varpi_i(s)}\frac{d \varpi_i^{\tau}(s)}{d\tau} \notag\\
    &
    \underset{(i)}{=} -\sum_{i,s}\frac{\partial {V}_i(\mu,\varpi^\tau) }{\partial \varpi_i(s)} \frac{d \varpi_i^\tau(s)}{dt} \notag\\
    &{{}\underset{(ii)}{=}\sum_{i,s}\frac{d^{\varpi^\tau}_{\mu}(s)}{\discount-1}\gamma_i(s){Q}_i(s;\varpi^\tau)^\top\lr{ \widetilde{\mathrm{br}}_i^{\tau,(\theta)}(s)-\varpi_i^\tau(s)},} \notag 
    \\
    &{ {}\underset{(iii)}{=}\sum_{i,s}\frac{d^{\varpi^\tau}_{\mu}(s)}{\discount-1}\gamma_i(s)(\Delta {Q}_i(s))^\top\lr{ \widetilde{\mathrm{br}}_i^{\tau,(\theta)}(s)-\varpi_i^\tau(s)}\notag} \\
    &{{}+ \sum_{i,s}\frac{d^{\varpi^\tau}_{\mu}(s)}{\discount-1}\gamma_i(s)({Q}_i(s;\varpi^\tau_i,\varpi_{-i}^{\tau,(\theta)}))^\top\lr{ \widetilde{\mathrm{br}}_i^{\tau,(\theta)}(s)-\varpi_i^\tau(s)}},
    \label{eq:one_stop}
    \end{align}
    {{}where \[\Delta {Q}_i(s) = {Q}_i(s;\varpi^\tau)-{Q}_i(s;\varpi^\tau_i,\varpi_{-i}^{\tau,(\theta)}).\]}Here, \((i)\) {is due to Lemma \ref{lem: Grad_equal}}, \((ii)\) is due to Lemma \ref{lem: TechnicalMain}(a) and \eqref{eq: policy_diff}, and {{}\((iii)\) is by adding and subtracting terms. Note that the first term in the RHS of \eqref{eq:one_stop} can be bounded as 
    \begin{align}
&\sum_{i,s}\frac{d^{\varpi^\tau}_{\mu}(s)}{\discount-1}\gamma_i(s)(\Delta {Q}_i(s))^\top\lr{ \widetilde{\mathrm{br}}_i^{\tau,(\theta)}(s)-\varpi_i^\tau(s)}\notag \\ 
       & {{}\leq \frac{1}{1-\discount}\sum_{i\in I}\max_{s\in S}\Big|(\Delta Q_i(s))^\top\lr{ \widetilde{\mathrm{br}}_i^{\tau, (\theta)}(s)-\varpi_i^\tau(s)}\Big|} \notag \\ 
        &\leq \frac{2}{1-\discount}\sum_{i\in I}\max_{s\in S,a_i\in A_i}\Big|(\Delta Q_i(s,a_i))\Big|\leq \frac{4|\playerSet|\sum_{i\in\playerSet}\theta_i u_{\max}}{(1-\discount)^3}, \label{eq: ExtraError}
    \end{align}
    where the first inequality is because \(\gamma_i(s)\leq 1\) as it denotes the fraction of time state \(s\) is visited, and last inequality is due to Lemma \ref{lem: TechnicalMain}(d).

Recall that \(\widetilde{\mathrm{br}}^{\tau,(\theta)}_i(s) \in {\mathrm{br}}_i(s;(\varpi_i^{\tau},\varpi_{-i}^{\tau,(\theta)}))\) where {{}$
{\mathrm{br}}_i(s;\varpi) 
    = \underset{{\hat{\varpi}_i\in\Delta(A_i)}}{\arg\max} ~ \hat{\varpi}_i^\top Q_i(s;\varpi)$} for every \(s\in S, \varpi\in \Pi\).  Therefore, \({Q}_i(s;(\varpi_i^{\tau},\varpi_{-i}^{\tau,(\theta)}))^\top\lr{ \widetilde{\mathrm{br}}_i^{\tau,(\theta)}(s)-\varpi_i^\tau(s)} \geq 0\). Additionally, given the fact that {{}\(\gamma_i(s)\geq \eta\)} for all \(i\in\playerSet,s\in\stateSet\),
 \eqref{eq:one_stop}, and \eqref{eq: ExtraError}, we obtain
\begin{align}\label{eq:two_stop}
    &\frac{d}{d\tau }\phi(\tau) \leq  \frac{4|\playerSet|\sum_{i\in I} \theta_i u_{\max}}{(1-\discount)^3} -\frac{\eta}{1-\discount}\sum_{i,s}d^{\varpi^\tau}_{\mu}(s)  \notag 
 \\ &\quad \cdot {Q}_i(s;(\varpi_i^\tau, \varpi_{-i}^{\tau,(\theta)}))^\top \lr{ \widetilde{\mathrm{br}}_i^{\tau,(\theta)}(s)-\varpi_i^\tau(s)}.  
    \end{align}

Additionally, let \(\BR_i \in \arg\max_{\pi_i\in \Pi_i}{}{V}_i(\mu,\policy_i,\varpi_{-i}^\tau)\) be a best response of player \(i\) if the joint strategy of other players is \(\varpi_{-i}^\tau\). Note that $\BR_i$ maximizes the total payoff instead of just maximizing the payoff of one-stage deviation. Therefore, $\BR_i$ can be different from the optimal one-stage deviation policy. We drop the dependence of $\BR_i$ on $\varpi_{-i}^\tau$ for notational simplicity.  

Recall that {{}\(D = \frac{1}{1-\discount} \max_{i,\varpi_{-i}}\|d^{\BR_i, \varpi_{-i}^\tau}_{\mu}/\mu\|_{\infty}\)}. We note that $D$ is finite under the assumption that $\mu$ has full support (Assumption \ref{as:basic}). Next, we provide an inequality that is crucial to bound \eqref{eq:two_stop}. Note that 
{{}\begin{align}
     &\sum_{i,s}d^{\BR_i, \varpi_{-i}^{\tau,(\theta)}}_{\mu}(s){Q}_i(s;\varpi_i^\tau, \varpi_{-i}^{\tau,(\theta)})^\top \lr{ \BR_i(s)-\varpi_i^\tau(s)}\notag \\ 
    &\leq \sum_{i,s}d^{\BR_i, \varpi_{-i}^{\tau,(\theta)}}_{\mu}(s)\notag \\
    &\qquad \cdot \max_{\hat{\pi}_i\in \Delta(A_i)}{Q}_i(s;\varpi_i^\tau, \varpi_{-i}^{\tau,(\theta)})^\top \lr{ \hat{\pi}_i(s)-\varpi_i^\tau(s)}\notag 
    \\
    &\leq \sum_{i,s}d_\mu^{\varpi^{\tau}}(s)\bigg\|\frac{d^{\BR_i, \varpi_{-i}^{\tau,(\theta)}}_{\mu}}{d_\mu^{\varpi^{\tau}}}\bigg\|_{\infty}\notag \\ &\quad \quad\quad \quad \cdot\max_{\hat{\pi}_i\in \Delta(A_i)}{Q}_i(s;\varpi_i^\tau, \varpi_{-i}^{\tau,(\theta)})^\top \lr{ \hat{\pi}_i(s)-\varpi_i^\tau(s)} \notag \\
    &\stackrel{\textit{(i)}}{\leq}   D \sum_{i,s}d_\mu^{\varpi^{\tau}}(s) \cdot\max_{\hat{\pi}_i\in \Delta(A_i)}{Q}_i(s;\varpi_i^\tau, \varpi_i^{\tau,(\theta)})^\top \lr{ \hat{\pi}_i(s)-\varpi_i^\tau(s)}
  \notag\\
    & = D \sum_{i,s}d_\mu^{\varpi^{\tau}}(s) \cdot{Q}_i(s;\varpi_i^\tau, \varpi_{-i}^{\tau,(\theta)})^\top \lr{ \widetilde{\mathrm{br}}_i^{\tau, (\theta)}(s)-\varpi_i^\tau(s)}, \notag
\end{align}}where 
\textit{(i)} is using\footnote{This is obtained by dropping all terms corresponding to \(k\geq 1\) in \eqref{eq: d_pi_eq}.} {{}\(d_{\mu}^{\varpi^{\tau,(\theta)}}(s)\geq (1-\discount)\mu(s)\)} along with the definition of \(D\). Therefore, 
\begin{equation}\label{eq:max}
\begin{split}
    &\sum_{i,s}d_\mu^{\varpi^{\tau}}(s) \cdot{Q}_i(s;\varpi_i^\tau, \varpi_{-i}^{\tau,(\theta)})^\top \lr{ \widetilde{\mathrm{br}}_i^{\tau, (\theta)}(s)-\varpi_i^\tau(s)} \\
    \geq &\frac{1}{D} \sum_{i,s}d^{\BR_i, \varpi_{-i}^{\tau,(\theta)}}_{\mu}(s){Q}_i(s;\varpi_i^\tau, \varpi_{-i}^{\tau,(\theta)})^\top \lr{ \BR_i(s)-\varpi_i^\tau(s)}.
    \end{split}
\end{equation}

{{}Then, from \eqref{eq:two_stop} and \eqref{eq:max}, we have
    
    \begin{align}\label{eq: PotentialFirst}
    &\frac{d}{dt}\phi(\tau) \leq  \frac{4|\playerSet|\sum_{i\in I}\theta_i u_{\max}}{(1-\discount)^3} -\frac{\eta}{D(1-\discount)}\sum_{i,s}d^{\BR_i, \varpi_{-i}^{\tau,(\theta)}}_{\mu}(s)\notag \\&\quad \quad \quad \quad \cdot {Q}_i(s;\varpi_i^\tau, \varpi_{-i}^{\tau,(\theta)})^\top \lr{ \BR_i(s)-\varpi_i^\tau(s)}. 
\end{align}}
Next, we note that
{\begin{align}\label{eq: PotentialSecond}
    &Q_i(s;\varpi_i^\tau, \varpi_{-i}^{\tau,(\theta)})^\top(\pi_i^\dagger(s) - \varpi_i^\tau(s)) \notag \\
\stackrel{(i)}{=}& \sum_{a_i\in A_i}{{}\lr{Q_i(s,a_i;\varpi_i^\tau, \varpi_{-i}^{\tau,(\theta)})- V_i(s,\varpi_i^\tau, \varpi_{-i}^{\tau,(\theta)})}}\notag \\ &\quad \cdot(\pi_i^\dagger(s,a_i) - \varpi_i^\tau(s,a_i)) \notag 
\\
\stackrel{(ii)}{=}&\sum_{a_i}\advFunc_i(s,a_i;\varpi_i^\tau, \varpi_{-i}^{\tau,(\theta)})\BR_i(s,a_i)
 = \advFunc_i(s,\BR_i;\varpi_i^\tau, \varpi_{-i}^{\tau,(\theta)}),
\end{align}
}where \((i)\) holds because \(V_i\) is not dependent on actions, and \((ii)\) follows from the definition of advantage function in \eqref{eq:advantageMain}. 
{{}Combining \eqref{eq: PotentialFirst} and \eqref{eq: PotentialSecond}, 
\begin{align}
    &\frac{d \phi(\tau)}{dt} \leq \frac{4|\playerSet|\sum_{i\in \playerSet}\theta_iu_{\max}}{(1-\discount)^3} \notag \\ &\quad \quad \quad -\frac{\eta}{D(1-\discount)}\sum_{i,s}d^{\BR_i, \varpi_{-i}^{\tau,(\theta)}}_{\mu}(s)\advFunc_i(s,\BR_i;\varpi_i^\tau,\varpi_{-i}^{\tau,(\theta)})  \notag \\ 
    &{=} \frac{4|\playerSet|\sum_{i\in \playerSet}\theta_iu_{\max}}{(1-\discount)^3} \notag \\ &\quad \quad \quad -({\eta}/{D}) \cdot \sum_{i}\lr{ {V}_i(\mu,\BR_i,\varpi_{-i}^{\tau, \theta})- {}{V}_i(\mu,\varpi_i^\tau, \varpi_{-i}^{\tau,\theta})} \label{eq: D_phi_final_second}
\end{align}
where the equality follows from the multi-agent performance difference lemma (Lemma \ref{lem: TechnicalMain}(b)). Next, note that 

\begin{align}
    &-({\eta}/{D}) \cdot \sum_{i}\lr{ {V}_i(\mu,\BR_i,\varpi_{-i}^{\tau, (\theta)})- {}{V}_i(\mu,\varpi_i^\tau, \varpi_{-i}^{\tau,(\theta)})} \notag \\
    &\leq -({\eta}/{D})\cdot  \sum_{i}\lr{ {V}_i(\mu,\BR_i,\varpi_{-i}^{\tau})- {}{V}_i(\mu,\varpi_i^\tau, \varpi_{-i}^{\tau})}\notag \notag  \\
&  {\quad + ({2\eta}/{D})\cdot {\max_{\pi_i\in\Pi_i}\sum_i|V_i(\mu,\pi_i,\varpi_{-i}^{\tau,(\theta)})-V_i(\mu,\pi_i,\varpi_{-i}^{\tau})|} \notag }  \\   
  &\leq -\frac{\eta}{D} \sum_{i}\lr{ {V}_i(\mu,\BR_i,\varpi_{-i}^{\tau})- {}{V}_i(\mu,\varpi)} + \frac{4\eta|\playerSet|\sum_{i\in I}\theta_i u_{\max}}{D(1-\discount)^2},\label{eq: V_diff_Equation}
\end{align}
where the first inequality is based on adding and subtracting the term \(-(\eta/D)\sum_i (V_i(\mu,\pi_i^\dagger,\varpi_{-i}^\tau)-V_i(\mu,\varpi_i^{\tau},\varpi_{-i}^\tau)\)), arranging terms and taking maximum over \(\pi_i\), and the last inequality is due to Lemma \ref{lem: TechnicalMain}(c). 

Combining \eqref{eq: D_phi_final_second} and \eqref{eq: V_diff_Equation}, we obtain 
\begin{align*}
 & {d\phi(\tau)}/{d\tau}\leq  \frac{4|\playerSet|\sum_{i\in I}\theta_i u_{\max}}{(1-\discount)^3} + \frac{4\eta |\playerSet|\sum_{i\in I}\theta_i u_{\max}}{D(1-\discount)^2}\\ & -(\eta/D)\cdot \sum_{i}\lr{ {V}_i(\mu,\BR_i,\varpi_{-i}^{\tau})- {}{V}_i(\mu,\varpi^\tau)}.  
\end{align*}
Since \(\pi_i^\dagger\) is a best response to \(\varpi^\tau_{-i}\), \( {}{V}_i(\mu,\BR_i, \varpi_{-i}^\tau) - {}{V}_i(\mu,\varpi^\tau)\geq 0\) for all \(i\). Furthermore, for any \(\varpi^{\tau}\not\in \textsf{NE}(\epsilon)\), there must exist at least one player \(i\in I\) and  best response policy \(\pi_i^{\dagger}\in \Pi_i\) such that ${V}_i(\mu,\pi_i^{\dagger},\varpi_{-i}^\tau) \geq {V}_i(\mu,\varpi^{\tau}_i,\varpi_{-i}^\tau)+ \epsilon$. 
Therefore, given that $\theta$ satisfies \eqref{eq:theta_constraint}, we have \(d\phi(\tau)/d\tau < 0\).}

\(\hfill\blacksquare\)  

\medskip 
Next, we present the following result that leverages Lemma \ref{lemma:d_potential} to obtain the convergent set of the sequence of policies induced by Algorithm \ref{alg:independent_decentralized}. 
\begin{lemma}\label{lemma: Policy_slow_continuous} Under the conditions stated in Theorem \ref{theorem:independent}, the sequence of policies \((\policy^t)_{t=0}^{\infty}\) given by Algorithm \ref{alg:independent_decentralized} almost surely converges to \(\Pi^\ast_{\epsilon}\) with probability 1. 
\end{lemma}
\begin{proof}
By applying the asynchronous stochastic approximation theory \cite[Theorem 4.7]{perkins2013asynchronous}, the asymptotic behavior of the policies \((\policy^t)_{t=0}^{\infty}\) is the same as the asymptotic behavior of \eqref{eq: policy_diff}. Therefore, it is sufficient to show that any absolutely continuous trajectory of \eqref{eq: policy_diff} will converge to \(\Pi^\ast_{\epsilon}\). 

From Lemma  \ref{lemma:d_potential}, we know that the potential function always (strictly) increases along the trajectory of \eqref{eq: policy_diff} outside \(\textsf{NE}(\epsilon)\). 
Since the potential function is bounded\footnote{The boundedness of potential function is without loss of generality as it is shift invariant.}, any absolutely continuous trajectory of \eqref{eq: policy_diff} will enter the set \(\textsf{NE}(\epsilon)\) in finite time\footnote{Even though we state that any absolutely continuous trajectory of \eqref{eq: policy_diff} enters \(\textsf{NE}(\epsilon)\), but it may leave that set and re-enter.}. Since \(\textsf{NE}(\epsilon)\subseteq \Pi^\ast_{\epsilon}\), once the trajectory of \eqref{eq: policy_diff} enters the set \(\textsf{NE}(\epsilon)\), it is already inside the set \(\Pi^\ast_{\epsilon}\). The proof concludes by showing that \(\Pi^\ast_{\epsilon}\) is an invariant set. 
Indeed, by noting that \(\Pi^\ast_{\epsilon}\) is a super-level set of the potential function and \(\Pi^\ast_{\epsilon}\) contains \(\textsf{NE}(\epsilon)\), we conclude that \(\Pi^\ast_{\epsilon}\) is an invariant set using Lemma \ref{lemma:d_potential}. Thus, any absolutely continuous trajectory of \eqref{eq: policy_diff} will enter the set \(\Pi^\ast_{\epsilon}\) and remain inside it forever.
\end{proof}
\smallskip

\medskip 

\textit{Proof of Theorem \ref{theorem:independent}.}
From Lemmas \ref{lemma:d_fast}-\ref{lemma: Policy_slow_continuous}, we conclude that the policy sequence \(\{\pi^{t}\}_{t=0}^{\infty}\) induced by Algorithm \ref{alg:independent_decentralized} converges to the set \(\Pi^\ast_{\epsilon}\) .

{
Next, we show that for any \(0\leq \epsilon\leq \epsilon'\), \(\Pi^\ast_{\epsilon}\subseteq \Pi^\ast_{\epsilon'}\). This follows from the fact that \(\textsf{NE}(\epsilon)\subseteq \textsf{NE}(\epsilon')\). Thus, \(\min_{\pi\in \textsf{NE}(\epsilon')}\Phi(\mu,\pi) \leq \min_{\pi\in \textsf{NE}(\epsilon)}\Phi(\mu,\pi).\) As a result, if \(\pi\in \Pi^\ast_{\epsilon}\), then \(\pi\in \Pi^\ast_{\epsilon'}\). 

Finally, we show that for every \(\epsilon\geq 0,\) \(\textsf{NE}(\epsilon) \subseteq \Pi^\ast_{\epsilon}.\) Suppose that \(\pi\in \textsf{NE}(\epsilon)\), then \(\Phi(\mu,\pi)\geq \min_{\pi'\in\textsf{NE}(\epsilon)}\Phi(\mu,\pi').\) This ensures that \(\pi\in\Pi^\ast_{\epsilon}\). 
}
}

\section{Numerical experiments}\label{sec:numerics}

In this section, we demonstrate the performance of the proposed learning dynamics (Algorithm \ref{alg:independent_decentralized}) in a Markov routing game (inspired by the example presented in \cite{leonardos2021global}). 
    Consider a parallel link network comprising of \(L\) links which is repeatedly used by \(\playerSet\) travelers (i.e. players). At every stage $k$, each player $i$ picks a link $a_i^k \in  [L]$ to commute.
The state of the network is \(s = (s_\ell)_{\ell\in[L]]} \), where \(s_\ell = 1\) represents that link \(\ell\) is unsafe and \(s_\ell=0\) represents that link $\ell$ is safe. 
The probability that a particular link becomes unsafe in the next time step $k+1$ is \(\lambda_{1}\) if the number of players using that link in stage $k$ is larger than or equal to a threshold \(\thresh\), and this probability is \(\lambda_{2}\) otherwise. 
 
 {Here, we consider the common interest rewards. That is,} given network state \(s\) and joint action \(a\), the utility of any player $i\in I$ is $u_i(s,a)= u(s,a)= \sum_{i\in I}\sum_{\ell=1}^{L}\mathbb{I}(a_i=\ell)\big(b_{\ell} - (1 + s_{\ell})m_{\ell}\sum_{j\in\playerSet}\mathbb{I}(a_{j} = \ell)\big)$. Here, \(b_\ell\) is the fixed utility of using link \(\ell \in [L]\), \(m_\ell\) is a link-dependent constant which weights the effect of congestion on the cost and  $(1 + s_{\ell})m_{\ell}\sum_{j\in\playerSet}\mathbb{I}(a_{j} = \ell)$ represents the cost of using $\ell$ given the total number of users on $\ell$ and the network state. 
The goal of every player \(i\in\playerSet\) is to choose a policy \(\policy_i:S\ra\Delta([L])\) to maximize the long run expected discounted payoff \(\avg[\sum_{k=0}^{\infty}\discount^ku(s^k,a^k)]\). {Due to the common reward structure the game is a Markov potential game.}
In this example, we set \(|\playerSet|=4\), \(L=2\), \(\thresh=2\), \(\lambda_{1} = 0.8\), \(\lambda_{2} = 0.2\), 
\(m_{1} = 2\), \(b_{1} = 9\), \(m_{2} = 4\), and \(b_{2} = 16\). We simulate for \(T = 10^{4}\) stages. The step size schedule \(\alpha_i(n) = 1/n^{0.5}\), \(\beta_i(n)=1/n\) for all \(i\in \playerSet\). The initial state of every link is sampled uniformly randomly. 

\begin{figure}[!ht]
        \centering
    \begin{subfigure}{0.5\textwidth}
    \centering
\includegraphics[width=\textwidth]{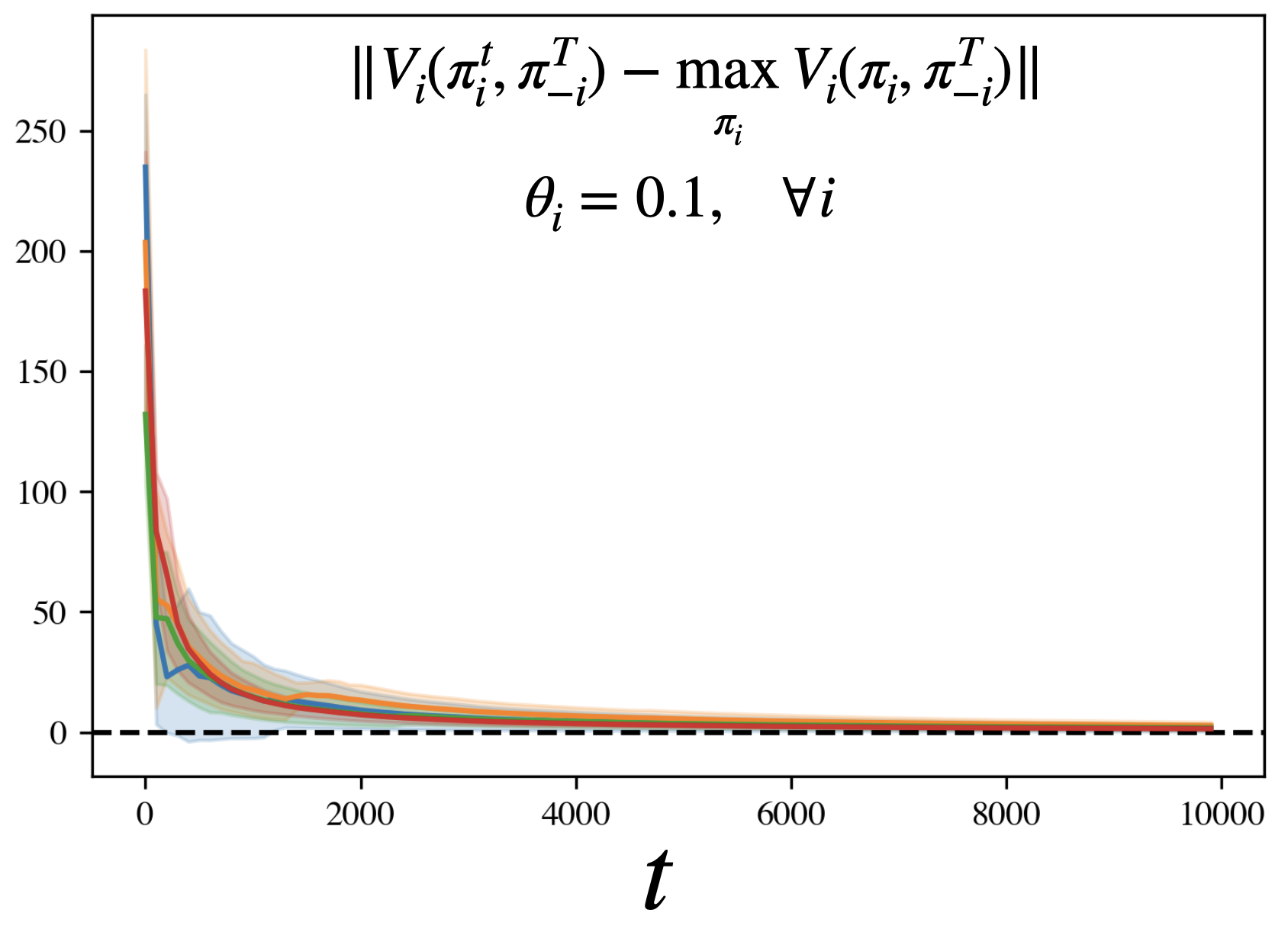}
    \caption{}
    \label{fig: F1}
    \end{subfigure}\\
    \begin{subfigure}{0.5\textwidth}
    \centering 
\includegraphics[width=\textwidth]{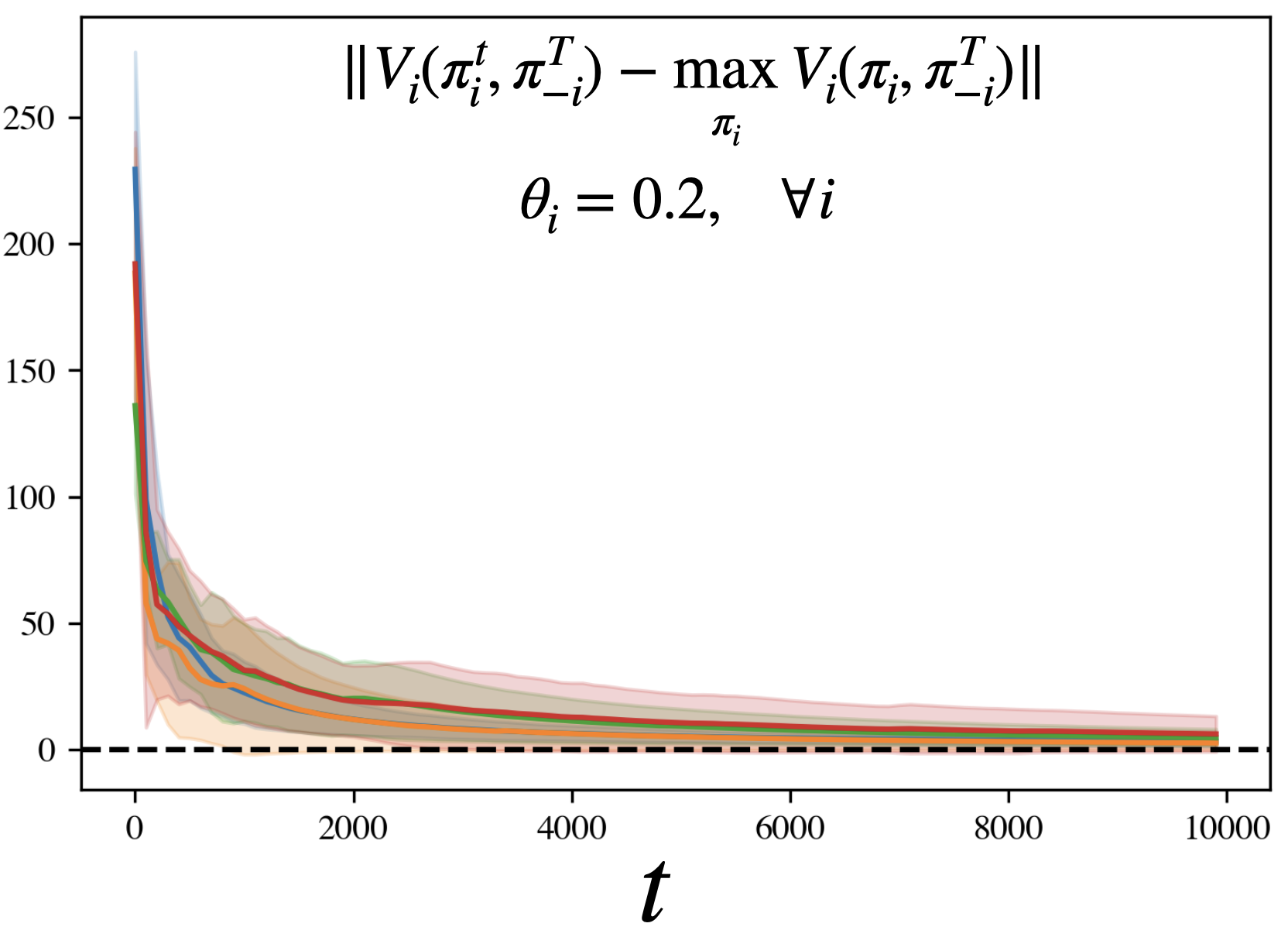}
    \caption{}
    \label{fig: F2}
    \end{subfigure}
    \caption{Variation of Nash  approximation gap during \(10^4\) steps of Algorithm \ref{alg:independent_decentralized}. The first (resp. second) figure shows the variation with exploration probability \(\theta_i = 0.1\) (resp. \(\theta_i = 0.2\)), for every \(i\in I.\) In each of the figures the four curves correspond to four players. 
    Each curve represents the mean value of the quantity over \(5\) trials, and we give error margins of \(\pm 1\) standard deviation.}
    \label{fig:ind_decentralized}
\end{figure}

To study the Nash approximation error with respect to the converged policy, we study the function: \(\|V_i(\cdot,\policy_i^{t},\policy_{-i}^{T}) - \max_{\policy_i'\in\Pi_i}V_i(\cdot,\policy_i',\policy_{-i}^T) \|_1\), where \(\policy_i^{T}\) is the converged policy update and \(\policy_i^t\) is \(k^{th}\) policy update. We observe that decreasing the exploration probabilities asymptotically leads to lower Nash approximation error (Figure \ref{fig:ind_decentralized}(a)-\ref{fig:ind_decentralized}(b)). 

\section{Conclusions}\label{sec: Conclusion}
{In this article, we analyzed the long-run behavior of a multi-agent decentralized reinforcement learning algorithm in infinite-horizon discounted Markov potential games. We demonstrated that when agents independently employ an actor-critic algorithm without communication, their learning processes converge to an approximate Nash equilibrium set.
There are several promising directions for future research. One avenue is to extend the analysis of decentralized learning dynamics beyond the framework of Markov potential games, exploring how these methods perform in more general multi-agent settings. Another important area is to investigate scenarios where agents adopt different learning algorithms. This could provide valuable insights into how the choice of learning dynamics influences the convergence behavior, long-term outcomes, and individual utilities of the agents in decentralized environments.}

\bibliography{refs}

\begin{thebibliography}{}

\bibitem[Alacaoglu et~al., 2022]{alacaoglu2022natural}
Alacaoglu, A., Viano, L., He, N., and Cevher, V. (2022).
\newblock A natural actor-critic framework for zero-sum {M}arkov games.
\newblock In {\em International Conference on Machine Learning}, pages
  307--366. PMLR.

\bibitem[Arslan and Y{\"u}ksel, 2016]{arslan2016decentralized}
Arslan, G. and Y{\"u}ksel, S. (2016).
\newblock Decentralized {Q}-learning for stochastic teams and games.
\newblock {\em IEEE Transactions on Automatic Control}, 62(4):1545--1558.

\bibitem[Bazzan, 2009]{bazzan2009opportunities}
Bazzan, A.~L. (2009).
\newblock Opportunities for multiagent systems and multiagent reinforcement
  learning in traffic control.
\newblock {\em Autonomous Agents and Multi-Agent Systems}, 18(3):342--375.

\bibitem[Bena{\"\i}m, 1999]{benaim1999dynamics}
Bena{\"\i}m, M. (1999).
\newblock Dynamics of stochastic approximation algorithms.
\newblock In {\em Seminaire de {P}robabilites XXXIII}, pages 1--68. Springer.

\bibitem[Bena{\"\i}m et~al., 2005]{benaim2005stochastic}
Bena{\"\i}m, M., Hofbauer, J., and Sorin, S. (2005).
\newblock Stochastic approximations and differential inclusions.
\newblock {\em SIAM Journal on Control and Optimization}, 44(1):328--348.

\bibitem[Borkar, 2002]{borkar2002reinforcement}
Borkar, V.~S. (2002).
\newblock Reinforcement learning in {M}arkovian evolutionary games.
\newblock {\em Advances in Complex Systems}, 5(01):55--72.

\bibitem[Borkar, 2009]{borkar2009stochastic}
Borkar, V.~S. (2009).
\newblock {\em Stochastic Approximation: A Dynamical Systems Viewpoint},
  volume~48.
\newblock Springer.

\bibitem[Brown and Sandholm, 2018]{brown2018superhuman}
Brown, N. and Sandholm, T. (2018).
\newblock Superhuman {AI} for heads-up no-limit poker: Libratus beats top
  professionals.
\newblock {\em Science}, 359(6374):418--424.

\bibitem[Cominetti et~al., 2010]{cominetti2010payoff}
Cominetti, R., Melo, E., and Sorin, S. (2010).
\newblock A payoff-based learning procedure and its application to traffic
  games.
\newblock {\em Games and {E}conomic {B}ehavior}, 70(1):71--83.

\bibitem[Daskalakis et~al., 2020]{daskalakis2020independent}
Daskalakis, C., Foster, D.~J., and Golowich, N. (2020).
\newblock Independent policy gradient methods for competitive reinforcement
  learning.
\newblock {\em Advances in neural information processing systems},
  33:5527--5540.

\bibitem[Ding et~al., 2022]{ding2022independent}
Ding, D., Wei, C.-Y., Zhang, K., and Jovanovic, M. (2022).
\newblock Independent policy gradient for large-scale {M}arkov potential games:
  Sharper rates, function approximation, and game-agnostic convergence.
\newblock In {\em International Conference on Machine Learning}, pages
  5166--5220. PMLR.

\bibitem[Fox et~al., 2022]{fox2022independent}
Fox, R., Mcaleer, S.~M., Overman, W., and Panageas, I. (2022).
\newblock Independent natural policy gradient always converges in {M}arkov
  potential games.
\newblock In {\em AISTATS}, pages 4414--4425. PMLR.

\bibitem[Fudenberg and Tirole, 1991]{fudenberg1991game}
Fudenberg, D. and Tirole, J. (1991).
\newblock {\em Game Theory}.
\newblock MIT press.

\bibitem[Guo et~al., 2021]{guo2021decentralized}
Guo, H., Fu, Z., Yang, Z., and Wang, Z. (2021).
\newblock Decentralized single-timescale actor-critic on zero-sum two-player
  stochastic games.
\newblock In {\em International Conference on Machine Learning}, pages
  3899--3909. PMLR.

\bibitem[Heliou et~al., 2017]{heliou2017learning}
Heliou, A., Cohen, J., and Mertikopoulos, P. (2017).
\newblock Learning with bandit feedback in potential games.
\newblock {\em Advances in Neural Information Processing Systems}, 30.

\bibitem[Hofbauer and Sandholm, 2002]{hofbauer2002global}
Hofbauer, J. and Sandholm, W.~H. (2002).
\newblock On the global convergence of stochastic fictitious play.
\newblock {\em Econometrica}, 70(6):2265--2294.

\bibitem[Hofbauer and Sigmund, 2003]{hofbauer2003evolutionary}
Hofbauer, J. and Sigmund, K. (2003).
\newblock Evolutionary game dynamics.
\newblock {\em Bulletin of the American {M}athematical {S}ociety},
  40(4):479--519.

\bibitem[Konda and Tsitsiklis, 1999]{konda1999actor}
Konda, V. and Tsitsiklis, J. (1999).
\newblock Actor-critic algorithms.
\newblock {\em Advances in Neural Information Processing Systems}, 12.

\bibitem[Krichene et~al., 2014]{krichene2014convergence}
Krichene, W., Drigh{\`e}s, B., and Bayen, A. (2014).
\newblock On the convergence of no-regret learning in selfish routing.
\newblock In {\em International Conference on Machine Learning}, pages
  163--171. PMLR.

\bibitem[Kutschinski et~al., 2003]{kutschinski2003learning}
Kutschinski, E., Uthmann, T., and Polani, D. (2003).
\newblock Learning competitive pricing strategies by multi-agent reinforcement
  learning.
\newblock {\em Journal of Economic Dynamics and Control}, 27(11-12):2207--2218.

\bibitem[Leonardos et~al., 2021]{leonardos2021global}
Leonardos, S., Overman, W., Panageas, I., and Piliouras, G. (2021).
\newblock Global convergence of multi-agent policy gradient in {M}arkov
  potential games.
\newblock {\em arXiv preprint arXiv:2106.01969}.

\bibitem[Leslie and Collins, 2006]{leslie2006generalised}
Leslie, D.~S. and Collins, E.~J. (2006).
\newblock Generalised weakened fictitious play.
\newblock {\em Games and Economic Behavior}, 56(2):285--298.

\bibitem[Leslie et~al., 2020]{leslie2020best}
Leslie, D.~S., Perkins, S., and Xu, Z. (2020).
\newblock Best-response dynamics in zero-sum stochastic games.
\newblock {\em Journal of Economic Theory}, 189:105095.

\bibitem[Macua et~al., 2018]{macua2018learning}
Macua, S.~V., Zazo, J., and Zazo, S. (2018).
\newblock Learning parametric closed-loop policies for {M}arkov potential
  games.
\newblock {\em arXiv preprint arXiv:1802.00899}.

\bibitem[Mao et~al., 2021]{mao2021decentralized}
Mao, W., Ba{\c{s}}ar, T., Yang, L.~F., and Zhang, K. (2021).
\newblock Decentralized cooperative multi-agent reinforcement learning with
  exploration.
\newblock {\em arXiv preprint arXiv:2110.05707}.

\bibitem[Marden, 2012]{marden2012state}
Marden, J.~R. (2012).
\newblock State based potential games.
\newblock {\em Automatica}, 48(12):3075--3088.

\bibitem[Marden et~al., 2009]{marden2009joint}
Marden, J.~R., Arslan, G., and Shamma, J.~S. (2009).
\newblock Joint strategy fictitious play with inertia for potential games.
\newblock {\em IEEE Transactions on Automatic Control}, 54(2):208--220.

\bibitem[Matignon et~al., 2012]{matignon2012independent}
Matignon, L., Laurent, G.~J., and Le~Fort-Piat, N. (2012).
\newblock Independent reinforcement learners in cooperative {M}arkov games: {A}
  survey regarding coordination problems.
\newblock {\em The Knowledge Engineering Review}, 27(1):1--31.

\bibitem[Mazumdar et~al., 2020]{mazumdar2019policy}
Mazumdar, E., Ratliff, L.~J., Jordan, M.~I., and Sastry, S.~S. (2020).
\newblock Policy-gradient algorithms have no guarantees of convergence in
  linear quadratic games.
\newblock In {\em Proceedings of the 19th International Conference on
  Autonomous Agents and MultiAgent Systems}, AAMAS '20, page 860–868.

\bibitem[Monderer and Shapley, 1996a]{monderer1996fictitious}
Monderer, D. and Shapley, L.~S. (1996a).
\newblock Fictitious play property for games with identical interests.
\newblock {\em Journal of {E}conomic {T}heory}, 68(1):258--265.

\bibitem[Monderer and Shapley, 1996b]{monderer1996potential}
Monderer, D. and Shapley, L.~S. (1996b).
\newblock Potential games.
\newblock {\em Games and {E}conomic {B}ehavior}, 14(1):124--143.

\bibitem[Narasimha et~al., 2022]{narasimha2022multi}
Narasimha, D., Lee, K., Kalathil, D., and Shakkottai, S. (2022).
\newblock Multi-agent learning via markov potential games in marketplaces for
  distributed energy resources.
\newblock In {\em 2022 IEEE 61st Conference on Decision and Control (CDC)},
  pages 6350--6357. IEEE.

\bibitem[Nuti et~al., 2011]{nuti2011algorithmic}
Nuti, G., Mirghaemi, M., Treleaven, P., and Yingsaeree, C. (2011).
\newblock Algorithmic trading.
\newblock {\em Computer}, 44(11):61--69.

\bibitem[Panageas and Piliouras, 2016]{panageas2016average}
Panageas, I. and Piliouras, G. (2016).
\newblock Average case performance of replicator dynamics in potential games
  via computing regions of attraction.
\newblock In {\em Proceedings of the 2016 ACM Conference on Economics and
  Computation}, pages 703--720.

\bibitem[Perkins and Leslie, 2013]{perkins2013asynchronous}
Perkins, S. and Leslie, D.~S. (2013).
\newblock Asynchronous stochastic approximation with differential inclusions.
\newblock {\em Stochastic Systems}, 2(2):409--446.

\bibitem[Perolat et~al., 2018]{perolat2018actor}
Perolat, J., Piot, B., and Pietquin, O. (2018).
\newblock Actor-critic fictitious play in simultaneous move multistage games.
\newblock In {\em International Conference on Artificial Intelligence and
  Statistics}, pages 919--928. PMLR.

\bibitem[Perolat et~al., 2015]{perolat2015approximate}
Perolat, J., Scherrer, B., Piot, B., and Pietquin, O. (2015).
\newblock Approximate dynamic programming for two-player zero-sum {M}arkov
  games.
\newblock In {\em International Conference on Machine Learning}, pages
  1321--1329. PMLR.

\bibitem[Prabuchandran et~al., 2014]{prabuchandran2014multi}
Prabuchandran, K., AN, H.~K., and Bhatnagar, S. (2014).
\newblock Multi-agent reinforcement learning for traffic signal control.
\newblock In {\em 17th International IEEE Conference on Intelligent
  Transportation Systems (ITSC)}, pages 2529--2534. IEEE.

\bibitem[Prasad et~al., 2015]{prasad2015two}
Prasad, H., LA, P., and Bhatnagar, S. (2015).
\newblock Two-timescale algorithms for learning nash equilibria in general-sum
  stochastic games.
\newblock In {\em Proceedings of the 2015 International Conference on
  Autonomous Agents and Multiagent Systems}, pages 1371--1379.

\bibitem[Sayin et~al., 2021]{sayin2021decentralized}
Sayin, M., Zhang, K., Leslie, D., Basar, T., and Ozdaglar, A. (2021).
\newblock Decentralized q-learning in zero-sum {M}arkov games.
\newblock {\em Advances in Neural Information Processing Systems}, 34.

\bibitem[Sayin et~al., 2022a]{sayin2020fictitious}
Sayin, M.~O., Parise, F., and Ozdaglar, A. (2022a).
\newblock Fictitious play in zero-sum stochastic games.
\newblock {\em SIAM Journal on Control and Optimization}, 60(4):2095--2114.

\bibitem[Sayin et~al., 2022b]{sayin2022fictitious}
Sayin, M.~O., Zhang, K., and Ozdaglar, A. (2022b).
\newblock Fictitious play in markov games with single controller.
\newblock In {\em Proceedings of the 23rd ACM Conference on Economics and
  Computation}, pages 919--936.

\bibitem[Shalev-Shwartz et~al., 2016]{shalev2016safe}
Shalev-Shwartz, S., Shammah, S., and Shashua, A. (2016).
\newblock Safe, multi-agent, reinforcement learning for autonomous driving.
\newblock {\em arXiv preprint arXiv:1610.03295}.

\bibitem[Song et~al., 2021]{song2021can}
Song, Z., Mei, S., and Bai, Y. (2021).
\newblock When can we learn general-sum {M}arkov games with a large number of
  players sample-efficiently?
\newblock {\em arXiv preprint arXiv:2110.04184}.

\bibitem[Swenson et~al., 2018]{swenson2018best}
Swenson, B., Murray, R., and Kar, S. (2018).
\newblock On best-response dynamics in potential games.
\newblock {\em SIAM Journal on Control and Optimization}, 56(4):2734--2767.

\bibitem[Tan, 1993]{tan1993multi}
Tan, M. (1993).
\newblock Multi-agent reinforcement learning: Independent vs. cooperative
  agents.
\newblock In {\em Proceedings of the 10th {I}nternational {C}onference on
  {M}achine {L}earning}, pages 330--337.

\bibitem[Tsitsiklis, 1994]{tsitsiklis1994asynchronous}
Tsitsiklis, J.~N. (1994).
\newblock Asynchronous stochastic approximation and {Q}-learning.
\newblock {\em Machine Learning}, 16(3):185--202.

\bibitem[Vinyals et~al., 2019]{vinyals2019grandmaster}
Vinyals, O., Babuschkin, I., Czarnecki, W.~M., Mathieu, M., Dudzik, A., Chung,
  J., Choi, D.~H., Powell, R., Ewalds, T., Georgiev, P., et~al. (2019).
\newblock Grandmaster level in {S}tarcraft {II} using multi-agent reinforcement
  learning.
\newblock {\em Nature}, 575(7782):350--354.

\bibitem[Yongacoglu et~al., 2023]{yongacoglu2023satisficing}
Yongacoglu, B., Arslan, G., and Y{\"u}ksel, S. (2023).
\newblock Satisficing paths and independent multiagent reinforcement learning
  in stochastic games.
\newblock {\em SIAM Journal on Mathematics of Data Science}, 5(3):745--773.

\bibitem[Zhang et~al., 2022]{zhangglobal}
Zhang, R., Mei, J., Dai, B., Schuurmans, D., and Li, N. (2022).
\newblock On the global convergence rates of decentralized softmax gradient
  play in {M}arkov potential games.
\newblock In {\em Advances in Neural Information Processing Systems}.

\bibitem[Zhang et~al., 2024]{zhang2021gradient}
Zhang, R., Ren, Z., and Li, N. (2024).
\newblock Gradient play in stochastic games: stationary points, convergence,
  and sample complexity.
\newblock {\em IEEE Transactions on Automatic Control}.

\end{thebibliography}
\bibliographystyle{apalike}

\newpage

\appendix
\section*{\center Appendix}
The appendix is organized as follows.  In Sec \ref{sec: PerkinsSA} we review the theory of two-timescale asynchronous stochastic approximation from \cite{perkins2013asynchronous}. In Sec \ref{app: ProofLemmaPD} we present the proofs of technical lemmas presented in Sec \ref{sec:result}.

\section{Review of Two-timescale asynchronous stochastic approximation}\label{sec: PerkinsSA}
In this section we review the results from  \cite{perkins2013asynchronous} on the theory of two-timescale asynchronous stochastic approximation. Note that we do not state their results in full generality but only to the extent necessary for this paper.

Let \(\{x^t\}_{t=1}^{\infty}, \{y^t\}_{t=1}^{\infty}\) be the stochastic approximation updates. Let \(x^t\in\R^{\dimXSA}\), \(y^t\in\R^{\dimYSA}\) for all \(t\in\{1,2,..\}\). Let \(\bar{\dimXSA}\subset [\dimXSA]\) (resp. \(\bar{\dimYSA}\subset [\dimYSA]\)) be the elements of \(x\) update (resp. \(y\) update) that have positive probability of being updated in the asynchronous update process. At iterate \(t\), let \(\bar{\dimXSA}^t\subset\bar{\dimXSA}\) and \(\bar{\dimYSA}^t\subset \bar{\dimYSA}\) be the elements that are updated. 
Let
\begin{align*}
    \tilde{n}^t(\compX)= \sum_{p=1}^{t}\mathbbm{1}(\compX\in\bar{\dimXSA}^p),\quad {n}^t(\compY) =
    \sum_{p=1}^{t}\mathbbm{1}(\compY\in\bar{\dimYSA}^p),
\end{align*}
for every \(\compX \in [\dimXSA]\) and \(\compY\in [\dimYSA]\).
Consider the following asynchronous stochastic approximation updates indexed by \(t\in\{1,2,..\}\)
\begin{equation} 
\begin{aligned}\label{eq: SA_Two_Init}
    x^t(\compX)&\in x^{t-1}(\compX)+\stepFast_{\compX}(\tilde{n}^t(\compX))\mathbbm{1}(\compX\in \bar{\dimXSA}^t)[F(\compX;x^{t-1},y^{t-1})\\&\quad +\martX^{t}({\compX})+d^t(\compX)],\quad\forall ~\compX\in [\dimXSA]\\ 
    y^t(\compY)&\in y^{t-1}(\compY)+\stepSlow_\compY({n}^t(\compY))\mathbbm{1}(\compY\in \bar{\dimYSA}^t)[G(\compY;x^{t-1},y^{t-1})\\&\quad +\martY^{t}({\compY})+e^t(\compY)],\quad \forall~ \compY\in[\dimYSA],
\end{aligned}
\end{equation}
where 
\begin{itemize}
\item[(i)] for any \(x\in\R^\dimXSA, y\in\R^{\dimYSA}\), \(F(x,y) = (F(\compX;x,y))_{\compX\in[\dimXSA]}\subset \R^{\dimXSA}\) and \(G(x,y)=(G(\compY;x,y))_{\compY\in[\dimYSA]}\subset\R^{\dimYSA}\)  are set-valued maps; 
\item[(ii)] \(\{\martX^t=(\martX^t(\compX))_{\compX\in [\dimXSA]}\},\{\martY^t=(\martY^t({\compY}))_{\compY\in [\dimYSA]}\}\) be martingale difference processes defined on \(\R^{\dimXSA},\R^{\dimYSA}\) respectively;
    \item[(iii)] \(\{d^t=(d^t(\compX))_{\compX\in\R^{\dimXSA}},e^t=(e^t(\compY))_{\compY\in\R^{\dimYSA}}\}\) are  asymptotically negligible error terms; 
    \item[(iv)] For every \(\compX\in [X], \compY\in [Y]\), \(\{\stepFast_\compX(n)\}_{n=0}^{\infty},\{\stepSlow_\compY(n)\}_{n=0}^{\infty}\) are the step sizes; 
    \item[(v)] \(x^0\in\R^{\dimXSA}, y^0\in\R^{\dimYSA}\) are initialized at some values. 
\end{itemize}

For every \(t\in \{1,2,...\}\), define 
\begin{align*}
     &\bar{\stepFast}^t= \max_{\compX\in \bar{\dimXSA}^t}\stepFast_\compX(\tilde{n}^t(\compX)), \quad   \mu^t(\compX) = \frac{\stepFast_\compX(\tilde{n}^t(\compX))}{\bar{\stepFast}^t} \mathbbm{1}(\compX\in\bar{\dimXSA}^t)\\
     &\bar{\stepSlow}^t = \max_{\compY\in \bar{\dimYSA}^t}\stepSlow_\compY(\tilde{n}^t(\compY)), \quad 
    \sigma^t(\compY) = \frac{\stepSlow_\compY(\tilde{n}^t(\compY))}{ \bar{\stepSlow}^t}\mathbbm{1}(\compY\in\bar{\dimYSA}^t) \\ 
     &\diagX^t = \textsf{diag}([\mu^t(1),\mu^t(2),...,\mu^t(\dimXSA)]),  \\&\diagY^t= \textsf{diag}([\sigma^t(1),\sigma^t(2),...,\sigma^t(\dimYSA)]).
\end{align*}
Using these notations we can concisely write \eqref{eq: SA_Two_Init} as 
\begin{equation}
\begin{aligned}\label{eq: SA_Condensed}
    x^t &\in x^{t-1}+ \bar{\stepFast}^t \diagX^t\lr{ F(x^{t-1},y^{t-1}) + \martX^t+d^t}\\
    y^t &\in y^{t-1}+\bar{\stepSlow}^t\diagY^t\lr{ G(x^{t-1},y^{t-1})+\martY^t+e^t }.
\end{aligned}
\end{equation}
We now state some assumption that are crucial to study the asymptotic property of the stochastic approximation \eqref{eq: SA_Condensed}. First, we introduce some important notations. 
Define \(\bar{H} \subset \bar{\dimXSA}\times\bar{\dimYSA}\) such that if \(i\in\bar{\dimXSA},j\in\bar{\dimYSA}\) then \((i,j)\in \bar{H}\) if and only if \(i,j\) have positive probability of occuring simultaneously. At iterate \(t,\) \(\bar{H}^t\in\bar{H}\) be the updated component in \([\dimXSA]\times[\dimYSA]\). Furthermore \(z^t=(x^t,y^t)\) be the joint update. Let \(\sigmaAlg^t=\sigma(\{\bar{H}^m\}_m,\{z^m\}_m,\{\tilde{n}^m(\compX)\},\{n^m(\compY)\}\}~\forall~m\leq t, \compX\in[\dimXSA],\compY\in[\dimYSA])\) be sigma-algebra containing all information upto iterate \(t\). For any positive integer \(K\) and a positive scalar \(\eta\), define \(\Omega^{\eta}_{K} = \{\textsf{diag}(\omega(1),...,\omega(K)): \omega(i)\in[\eta,1] ~\forall i=1,2,..,K\}\). 

Next, we present the assumptions required in \cite{perkins2013asynchronous} to study the asymptotic behavior of two-timescale asynchronous stochastic approximation update \eqref{eq: SA_Two_Init}.
\begin{assumption}\label{assm: PerkinsAssumptions}
Let the following assumptions hold 
\begin{itemize}
    \item[(A1)]\label{assm: A1} For compact sets \(\setX\subset\R^\dimXSA,\setY\subset\R^\dimYSA\), \(x^t\in \setX, y^t\in \setY\) for all \(t\in\{0,1,...\}\).
    \item[(A2)]\label{assm: A2} \(\{d^t\},\{e^t\}\) are bounded sequence such that \(\lim_{t\ra\infty}d^t=\lim_{t\ra\infty} e^t=0\).
    \item[(A3)]\label{assm: A3} Following must be true for the stepsizes:
    \begin{itemize}
         \item[(i)] For every \(\compX\in[X],\compY\in [Y]\), \(\sum_{n}\stepFast_\compX(n)=\infty, \sum_n\stepSlow_\compY(n)=\infty\), \(\lim_{n\ra\infty}\stepFast_\compX(n)=\lim_{n\ra\infty}\stepSlow_\compY(n)=0\) and \(\{\stepFast_\compX(n)\},\{\stepSlow_\compY(n)\}\) are non-increasing sequences.
    \item[(ii)] For any \(\lambda\in(0,1)\), \(\compX\in [X]\) and \(\compY\in [Y]\) it holds that \(\sup_n \stepFast_\compX([\lambda n])/\stepFast_\compX(n) < A_\lambda < \infty\), \(\sup_n \stepSlow_\compY([\lambda n])/\stepSlow_\compY(n) < A_\lambda < \infty\).
    \item[(iii)] For every \(\compX\in [X], \compY\in [Y]\) it holds that \(\lim_{n\ra\infty}\stepSlow_\compY(n)/\stepFast_\compX(n) = 0\).
    \item[(iv)] For every \(\compX,\compX'\in [X]\), \(\compY,\compY'\in [Y]\), there exists  \(0<\xi_{\compX\compX'}^\alpha<  \zeta_{\compX\compX'}^\alpha <\infty\), and \(0<\xi_{\compY\compY'}^\beta< \zeta_{\compY\compY'}^\beta<\infty\) such that  \(\frac{\stepFast_\compX(n)}{\stepFast_{\compX'}(n)}\in [\xi_{\compX\compX'}^\alpha, \zeta_{\compX\compX'}^\alpha]\) and \(\frac{\stepSlow_\compY(n)}{\stepSlow_{\compY'}(n)} \in [\xi_{\compY\compY'}^\beta, \zeta_{\compY\compY'}^\beta]\) for all $n$.
    \end{itemize}
    \item[(A4)]\label{assm: A4} The maps \(F(\cdot,\cdot), G(\cdot,\cdot)\)  are such that 
    \begin{itemize}
    \item[(i)] \(F:\setX\times \setY \rightrightarrows \setY\) is upper semi-continuous, for every \(z\in\setX\times \setY  \), \(F(z)\) is non-empty, compact, convex subset of \(\setY\), and \(\sup_{t\in F(z)}\|t\|\leq c(1+\|z\|)\) where \(c\) is a constant independent of \(z\).
        \item[(ii)] \(G:\setX\times \setY\rightrightarrows \setX\) is upper semi-continuous. \(G(x, \cdot)\) is non-empty, convex and compact and satisfy
        \(\sup_{t\in G(x,y)}\|t\|\leq c(1+\|y\|)\) where \(c\) is a constant independent of \(x,y\).
    \end{itemize}
    \item[(A5)]\label{assm: A5} 
    \begin{itemize}
        \item[(i)] for all \(z\in \setX\times \setY\) and \(h^{t-1},h^t\in \bar{H}\), 
        \begin{align*}
            &\Pr\lr{ \bar{H}^t=h^t|\mathcal{F}^{t-1} } \\ &\quad = \Pr(\bar{H}^t=h^t|\bar{H}^{t-1}=h^{t-1},z^{t-1}=z)
        \end{align*}
        \item[(ii)] For any \(z\in \setX\times \setY\) the transition probability
       \begin{align}\label{eq: TransitionReview}
     &\mathcal{P}(z;h^t,h^{t-1}) \notag\\&\quad \defas\Pr(\bar{H}^t=h^t|\bar{H}^{t-1}=h^{t-1},z^{t-1}=z)
       \end{align} 
       form aperiodic and irreducible Markov chain over \(\bar{H}\) and for every \(\compX\in \dimXSA\) and \(\compY\in \dimYSA\) there exists \(h,h'\in\bar{H}\) such that \(\compX\in h\) and \(\compY\in h'\).     
        \item[(iii)] the map \(z\mapsto \mathcal{P}(z;{h^t,h^{t-1}})\) is Lipschitz.
    \end{itemize}
    \item[(A6)]\label{assm: A6} 
    For some \(q\geq 2\), \(\sum_n \stepFast_\compX(n)^{1+q/2} < \infty\) and \(\sup_t\avg\ls{\|\martX^t\|^{q}}<\infty\) for every \(\compX\in[X]\).  
    For some \(q'\geq 2\), \(\sum_n \stepSlow_\compY(n)^{1+q'/2} < \infty\) and \(\sup_t\avg\ls{\|\martY^t\|^{q}}<\infty\) for every \(\compY\in [Y]\).
\item[(A7)]\label{assm: A7} For all \(y\in \setY\) and every \(\phi>0\) the differential inclusion 
\begin{align*}
    \frac{d}{d\tau}{x}^\tau \in \Omega^{\phi}_{\dimXSA}\cdot F(x^\tau,y),
\end{align*}
    has unique global attractor \(\Lambda(y)\), where \(\Lambda:\R^\dimYSA\ra\R^\dimXSA\) is bounded, continuous and single-valued for all \(y\in \setY\).
\end{itemize}
\end{assumption}

\begin{theorem}[Fast-timescale convergence]\label{thm: Fastconvergence} \cite[Corollary 4.4]{perkins2013asynchronous}
Under assumption (A1)-(A7) in Assumption \ref{assm: PerkinsAssumptions}, with probability 1,
\begin{align*}
    (x^t,y^t) \ra \{ (\Lambda(y),y): y\in \setY \} \quad \text{as} \quad t\ra\infty.
\end{align*} 
\end{theorem}

\medskip

Next, we present the corresponding convergence results for the slow updates, \(\{y^{t}\}\). Prior to that, we define the linearly interpolated trajectory  \(\{\bar{y}^\tau\}_{\tau\in \mathbb{R}_+}\) defined as 
\begin{align*}
    \bar{y}^{\bar{\tau}^t+s} = y^t + s\frac{y^{t+1}-y^t}{\bar{\beta}^{t+1}}, \quad s\in [0,\bar{\beta}^{t+1}),
\end{align*}
where \(\bar{\tau}^t = \sum_{p=0}^{t}\bar{\beta}^t\).

Define \(G^{\Lambda}:\R^\dimYSA\ra\R^\dimYSA\) as \(G^{\Lambda}(y)=G(\Lambda(y),y)\). {Furthermore, let \(\bar{G}^{\Lambda,\eta}(y) = \Omega^{\eta}_{\dimYSA} G^{\Lambda}(y)\), where \(\eta = \kappa/A_{\kappa}\) for some \(0 < \kappa \leq \min_{z\in \tilde{\mathcal{S}}\times \mathcal{S}, \compY\in [Y]}\psi_{z}(\compY)\), and \(\psi_{z}\in \Delta(Y)\) is the marginal of the stationary distribution of the Markov chain on \(\bar{H}\) with transition kernel \(\mathcal{P}(z;h,h')\) \cite[Lemma A.1, Appendix A.3]{perkins2013asynchronous}.} Consider the following differential inclusion 
\begin{align}\label{eq: app_slow_appendix}
    \dot{y}^\tau \in \bar{G}^{\Lambda,\eta}(y^\tau).
\end{align}

\begin{theorem}[Slow-timescale convergence] {\cite[Theorem 4.7]{perkins2013asynchronous}} \label{thm: SlowConvergence}
If the conditions (A1)-(A7) in Assumption \ref{assm: PerkinsAssumptions} are satisfied then
\(\{\bar{y}^\tau\}_{\tau\in \mathbb{R}_+}\) is an asymptotic pseudo-trajectory to \eqref{eq: app_slow_appendix}. 
\end{theorem}

\begin{remark}
Note that \cite{perkins2013asynchronous} assume that for every \(\compX,\compX'\in [X], \compY,\compY'\in[Y]\) it holds that \(\stepFast_{\compX}(\cdot) = \stepFast_{\compX'}(\cdot)\) and \(\stepSlow_{\compY}(\cdot) = \stepSlow_{\compY'}(\cdot)\). However, they easily generalize under the setting of heterogeneous step 
sizes considered here due to Assumption (A3)-(iv). Indeed, Theorem \ref{thm: Fastconvergence} (resp. Theorem \ref{thm: SlowConvergence}) follow similar to \cite{perkins2013asynchronous}  if we fix a \(\tilde{\compX}\in [X]\) 
(resp. \(\tilde{\compY}\in [Y]\)) and bound the relative evolution of step sizes at fast (resp. slow) timescale \(\compX\neq \tilde{\compX}\) (resp. \(\compY\neq \tilde{\compY}\) with respect to \(\tilde{\compX}, \tilde{\compY}\) using Assumption (A3)-(iv). 
\end{remark}\newpage
\section{Remaining Proofs}\label{app: ProofLemmaPD}
For clear presentation, we define \({Q}_i(s,\policy_i';\policy)=\policy'_i(s)^\top {Q}_i(s;\policy)\). 
{Recall, for any \(\pi\in \Pi, \theta\in (0,1)^{|I|}\), we define \(\pi^{(\theta)}\in \Pi\) such that for every \(s\in S, i\in I\), {\(\pi_{i}^{(\theta)}(s):= (1-\theta_i)\pi_{i}(s)+\theta_i(1/|A_{i}|)\cdot \mathbbm{1}_{A_{i}}\) to be a perturbed version of policy \(\pi\) due to exploration parameter \(\theta\).}}

\subsection{Proof of Lemma \ref{lemma:d_fast}}\label{ssec: d_fast_lemma}
The proof follows by verifying that Assumption \ref{assm: PerkinsAssumptions} (A1)-(A7) are satisfied and then evoking Theorem \ref{thm: Fastconvergence}. Towards that goal, we first verify Assumption \ref{assm: PerkinsAssumptions} (A1)-(A7).

 Before verifying the conditions for two-timescale asynchronous stochastic approximation stated in Section \ref{sec: PerkinsSA}, we introduce some notations. 
For any \(\policy\in\Pi\), \(i\in\playerSet,a_i\in\Ai,s\in\stateSet\), we define
\begin{align}\label{eq: Bellman}
    &\bellman_i^\policy \localQTilde_i(s,a_i) \defas \stagePayoff_i(s,a_i,\policy_{-i})\notag \\&\quad +\discount \sum_{s'}P(s'|s,a_i,\policy_{-i})\policy_i(s')^\top \localQTilde_i(s'),
\end{align}
which is analogous to Bellman operator in the setup of this paper. 
Furthermore, for any \(\pi\in \Pi\), we define
\begin{align}\label{eq: BellmanHat}
&{\bellmanHat_i^\policy \localQTilde_i(s,a_i)\defas  \stagePayoff_i(s,a_i,a_{-i})
+\discount \policy_i(s')^\top \localQTilde_i(s'),}
\end{align}
where \(a_{-i}\sim\policy_{-i}(s)\) and \(s'\sim 
P(\cdot|s,a_i,\pi_{-i})\). Moreover, for any \(s\in\stateSet,i\in\playerSet,a_i\in\Ai,\) we define
\begin{align}\label{eq: BrisAlg}
    \bar{\mathrm{br}}{}_i(s;q) = \underset{\pi\in \Delta(A_i)}{\arg\max}~\pi^\top q_i(s),
\end{align}
where for every \(i\in I, s\in S\), \(q_i(s)\in \R^{|A_i|}\).
Using the above notations, we re-write \eqref{eq:d_q}-\eqref{eq:d_pi} as 

\begin{subequations}
{ \begin{align}\label{eq: BellmanAlg}
    &\localQTilde_i^{t}(s,a_i) \notag\\
    =& \localQTilde_i^{t-1}(s,a_i) + \stepFast_i(\tilde{n}^t_i(s,a_i))\mathbbm{1}\{(s,a_i)=(s^{t-1},a^{t-1}_i)\}\notag\\&\cdot( \bellman_i^{(\policy^{t-1}_i,\policy^{t-1, (\theta)}_{-i})}\localQTilde_i^{t-1}(s,a_i)-\localQTilde_i^{t-1}(s,a_i)+\martX^t_i(s,a_i)),
    \end{align}}
    \begin{align} \label{eq: BellmanAlgPi}
    &\policy_i^{t}(s)\in \policy_i^{t-1}(s)+ \beta_i(\nk(s))\mathbbm{1}\{s=s^{t-1}\} \cdot \notag \\&\quad \cdot \lr{\bar{\mathrm{br}}{}_i(s;q^{t-1}) - \policy_i^{t-1}(s)}, 
\end{align}
\end{subequations}
for all \((s,a_i)\in \stateSet\times \Ai\), \(i\in\playerSet\), where 

{\begin{align}\label{eq: Mart}
    \martX^t_i(s,a_i)&= \bellmanHat_i^{(\policy^{t-1}_i, \policy^{t-1,(\theta)}_{-i})}\localQTilde_i^{t-1}(s,a_i) \notag \\ &\quad \quad -\bellman_i^{(\policy^{t-1}_i, \policy^{t-1,(\theta)}_{-i})}\localQTilde_i^{t-1}(s,a_i).
\end{align}}

Note that \(\avg[\martX^t_i(s,a_i)|\mathcal{F}^{t-1}] = 0\) where \(\mathcal{F}^{t-1}=\sigma(\{(s^m,a^m)\}_m,\{\localQTilde_i^m\}_{m},\{\policy_i^m\}_{m}: m\leq t-1, i\in \playerSet)\) is the sigma-algebra comprising of history till stage \(t-1\). Consequently, \(\{\martX^t_i\}\) is a martingale difference sequence. The updates \eqref{eq: BellmanAlg}-\eqref{eq: BellmanAlgPi} are now cast in the same formulation as in \eqref{eq: SA_Two_Init}. The asynchronous q-estimate updates  \eqref{eq: BellmanAlg} and the policy updates \eqref{eq: BellmanAlgPi} both have \(|\Pi_{i=1}^{\playerSet} (\stateSet\times \Ai)|\) components.

We now verify Assumption \ref{assm: PerkinsAssumptions} (A1)-(A7) one by one 

\noindent (i) First, we show that (A1) in Assumption \ref{assm: PerkinsAssumptions} is satisfied with \((\localQTilde^t,\policy^t)\) update \eqref{eq: BellmanAlg}-\eqref{eq: BellmanAlgPi}. Let \(\bar{
    \stagePayoff} = \max_{i,s,a}|\stagePayoff_i(s,a)|\). 
Moreover let \(\bar{R}=\max\{\bar{\stagePayoff}/(1-\discount),\max_i\|\localQTilde_i^0\|_{\infty}\}\). Then we claim that \(\|\localQTilde_i^t\|_{\infty}\leq \bar{R}\) for all \(t=\{0,1,2,..\}\). We show this by induction. It holds for \(t=0\) by construction. Suppose it holds for \(t=m-1\) for some \(m\) then we show that it also holds for \(t=m\). Indeed, we note from \eqref{eq: BellmanAlg} that \(\localQTilde_i^{t}\) is a convex combination\footnote{This is because we assume that \(\alpha(n)\in(0,1)\) in Assumption \ref{as:stepsize}.} of \(\localQTilde_i^{t-1}\) and \( \bellman_i^{(\policy^{t-1}_i,\policy^{t-1,(\theta)}_{-i})}\localQTilde_i^{t-1}(s,a_i)+\martX^t_i(s,a_i) \). Using \eqref{eq: BellmanHat} and \eqref{eq: Mart}  we see that
\begin{align*}
        &{\|\bellman_i^{(\policy^{t-1}_i,\policy^{t-1, (\theta)}_{-i})}\localQTilde_i^{t-1}+\martX^t_i\|_{\infty} }\\ &= \|\bellmanHat_i^{(\policy^{t-1}_i,\policy^{t-1, (\theta)}_{-i})}\localQTilde^{t-1}_i\|_{\infty} \\ &\leq \bar{u}+\delta \bar{R}   \leq (1-\discount)\bar{R}+\discount \bar{R} = \bar{R}.
    \end{align*}
    This shows that \(\|\localQTilde_i^t\|_{\infty}\leq \bar{R}\). Moreover note that \(\policy^{t}\in \Pi\) which is product simplex and is always compact.
    
    \noindent (ii) {Since we do not have any asymptotically negligible error terms in the asynchronous updates, Assumption\ref{assm: PerkinsAssumptions}-(A2) is immediately satisfied}
    
    \noindent (iii) Next we note Assumption \ref{assm: PerkinsAssumptions}-(A3) is satisfied due to Assumption \ref{as:stepsize}.

    \noindent (iv) Now we show Assumption \ref{assm: PerkinsAssumptions}-(A4) is satisfied. First, we concisely write the mean fields of \eqref{eq: BellmanAlg}-\eqref{eq: BellmanAlgPi} as follows 
    {\begin{align*}
       & F((s,a_i);\localQTilde,\policy,\theta) \defas \bellman^{(\policy_i,\policy^{(\theta)}_{-i})}_i\localQTilde_i(s,a_i) - \localQTilde_i(s,a_i), \\ 
        &G((s,a_i);\localQTilde,\policy) = \bar{\mathrm{br}}{}_i(s, a_i;q) - \policy_i(s,a_i),
    \end{align*}
    for every \(s\in\stateSet\), \(i\in\playerSet\), \(a_i\in\Ai\). Define \(F(\localQTilde,\policy,\theta) = \lr{F((s,a_i);\localQTilde,\policy, \theta)}_{s\in\stateSet,i\in\playerSet,a_i\in A_i}\), \(G(\localQTilde,\policy) = \lr{G((s,a_i);\localQTilde,\policy)}_{s\in\stateSet,i\in\playerSet,a_i\in A_i}\). We note that both \(F,G\) are continuous as demanded in Assumption \ref{assm: PerkinsAssumptions}-(A4). Furthermore, observe that
    \begin{align*}
    \|F(\localQTilde,\policy,\theta)\|_{\infty} &\leq \|\bellman^{(\policy_i, \policy^{(\theta)}_{-i})}_i\localQTilde_i\|_{\infty}+\|\localQTilde\|_{\infty}\\&\leq \bar{u}+\delta\|\localQTilde\|_{\infty}+\|\localQTilde\|_{\infty}\leq \tilde{c}(1+\|\localQTilde\|_{\infty}),
    \end{align*}
    where \(\tilde{c} = \max\{\bar{u},1+\delta\}\). Also note that
    \(
    \sup_{w\in G(\localQTilde,\policy)}\|w\|_{\infty} \leq 1+\|\policy\|_{\infty}.
    \)
    Thus we conclude that Assumption \ref{assm: PerkinsAssumptions}-(A4) is satisfied.}
    
    \noindent (v) We now verity Assumption \ref{assm: PerkinsAssumptions}-(A5). Consider \(h, h'\in \bar{H}\) such that \(h=((s,a_1),(s,a_2),..(s,a_\playerSet))\) and \(h'=((s',a_1'),(s,a_2'),...(s',a_\playerSet'))\).  Moreover, let \(z = (\localQTilde,\policy)\) then,
\begin{align}\label{eq: TransitionIndex}
      \mathcal{P}(z;h,{h}') = \transition(s'|s,a)\prod_{i\in\playerSet}\policy_i^{(\theta)}(s',a_i'),
    \end{align}
    where \(a=(a_i)_{i\in\playerSet}\) and the function \(\mathcal{P}(z;h,{h}')\) is defined in \eqref{eq: TransitionReview}. Since \(\theta>0\), all actions have positive probability of being selected. That is, for every \(k\in\mathbb{N}\) we have \(\policy_i^{t,(\theta)}(s,a_i)>\theta/|A_i|\) for all \(s\in\stateSet,i\in\playerSet,a_i\in\Ai\). Moreover, we impose Assumption \ref{as:basic} on transition matrix which ensures that every state is visited with some non-zero probability. Thus, Assumption \ref{assm: A5}(A5)-(i) and \ref{assm: A5}(A5)-(ii) are satisfied. Finally Assumption \ref{assm: A5}(A5)-(iii) is satisfied by noting that \eqref{eq: TransitionIndex} is Lipschitz in \(\policy\) and therefore in \(z\).
    
    \noindent (vi) Assumption \ref{assm: PerkinsAssumptions}-(A6) is satisfied by noting that (a) \(\martX\) is a bounded martingale difference sequence and (b) the step size condition in Assumption \ref{as:stepsize}-(ii) holds. 
    
    \noindent (vii) {For any \(\phi>0\), \(\policy\in\Pi\) consider the differential equation 
    \begin{align}\label{eq: BellmanContraction}
      \frac{d}{d\tau}{\localQTilde}_i^\tau = \Omega_{A_i}^{\phi}\lr{ \bellman_i^{(\policy_i,\policy^{(\theta)}_{-i})}\localQTilde_i^\tau  -\localQTilde_i^\tau }, \quad \forall \ i\in \playerSet,
    \end{align}} where \(\Omega^{\phi}_{\Ai} = \{\textsf{diag}(\omega(1),...,\omega(|\Ai|)): \omega(k)\in[\phi,1] ~\forall k=1,2,..,|\Ai|\}\).   In order to verify Assumption \ref{assm: PerkinsAssumptions}-(A7), we show that \eqref{eq: BellmanContraction} has unique global attractor for every \(\policy\in\Pi\).

    We show that \(\bellman_i^{\policy}\) is a contraction for every \(\policy\in\Pi\). For any $q, \bar{q}$, 
    \begin{align*}
        &\bellman^{\policy}_i q_i(s,a_i) - \bellman^{\policy}_i \bar{q}_i(s,a_i)\\
        =& \discount\sum_{s'}P(s'|s,a_i,\pi_{-i})\sum_{a_i'\in\Ai}\policy_i(s',a_i')\lr{ q_i(s',a_i')-\bar{q}_i(s',a_i') }.  
    \end{align*}
    Thus, for every \(s\in\stateSet,i\in\playerSet,a_i\in\Ai\), we have 
    \(
|\bellman_i^{\policy}q_i(s,a_i)-\bellman_i^{\policy}\bar{q}_i(s,a_i)|\leq \discount \|q_i-\bar{q}_i\|_{\infty}.
    \)
    Consequently, \(
\|\bellman_i^{\policy}q_i-\bellman_i^{\policy}\bar{q}_i\|_{\infty}\leq \discount \|q_i-\bar{q}_i\|_{\infty}
    \), and \(\bellman_i^{\policy}\) is a contraction mapping. 
    
   {Consequently, \(\bellman_i^{(\policy_i,\policy^{(\theta)}_{-i})}\) is a contraction and
    \eqref{eq: BellmanContraction} has a unique global attractor, which is the fixed point of the mapping \(\bellman_i^{(\policy_i,\policy^{(\theta)}_{-i})}\) \cite[Chapter 7.4]{borkar2009stochastic}. Moreover, from the definition it follows that \({Q}_i(\cdot,\cdot ;(\policy_i, \policy^{(\theta)}_{-i}))\)  is the fixed point of \(\bellman_i^{(\policy_i,\policy^{(\theta)}_{-i})}\). That is, for every \(s\in\stateSet,i\in\playerSet,a_i
        \in \Ai\), \(\bellman_i^{(\policy_i, \policy^{(\theta)}_{-i})}{Q}_i(s,a_i;(\policy_i, \policy^{(\theta)}_{-i})) = {Q}_i(s,a_i;(\policy_i, \policy^{(\theta)}_{-i})).\) }
    Additionally, \(\|{Q}_i(\cdot,\cdot;(\policy_i, \policy^{(\theta)}_{-i}))\|_{\infty} \leq \frac{u_{\max}}{1-\discount}\), and \({Q}_i(\cdot,\cdot;\policy )\) is also continuous in \(\policy\). 
Thus, Assumption \ref{assm: PerkinsAssumptions}-(A7) is satisfied.
Finally, the claim in Lemma \ref{lemma:d_fast} follows by Theorem \ref{thm: Fastconvergence}. 
 \subsection{Proof of Lemma \ref{lem: TechnicalMain}}\label{ssec: LemmaTechnicalMain}
 We prove (a)-(c) in sequence 
 
\noindent(a)
We claim that for any integer \(K\geq 0\), \(\mu\in\Delta(S), \pi\in \Pi, s\in\stateSet,i\in\playerSet,a_i\in\Ai\), 
\begin{align}\label{eq: InductionPG}
    \frac{\partial {V}_i(\mu,\policy)}{\partial\policy_i(s,a_i)}&=\avg\ls{\sum_{k=0}^{K}\discount^k \mathbbm{1}(s^k=s)} {Q}_i({s},a_i;\policy) \notag\\&\hspace{2cm}+ \delta^{K+1}\avg\ls{ \frac{\partial {V}_i(s^{K+1},\policy)}{\partial \policy_i(s,a_i)}},
\end{align}
where \(s_0\sim\mu, a^{k-1}\sim\policy(s^{k-1}), s^k\sim P(\cdot|s^{k-1},a^{k-1})\).
We prove this claim by induction. Indeed, this holds for \(K=0\) by noting that 
{\small\begin{align*}
    &\frac{\partial {V}_i(\mu,\policy)}{\partial\policy_i(s,a_i)} = \frac{\partial}{\partial \policy_i(s,a_i)}\sum_{\bar{s} \in S}\mu(\bar{s})\lr{ \sum_{\bar a_i\in\Ai}\policy_i(\bar{s},\bar a_i)  {Q}_i(\bar{s},\bar a_i;\policy)}\\
    =& \frac{\partial}{\partial \policy_i(s,a_i)}\bigg(\sum_{\bar{s}}\mu(\bar{s})\sum_{\bar{a}_i}\policy_i(\bar{s},\bar{a}_i)\bigg(\stagePayoff_i(\bar{s},\bar{a}_i,\policy_{-i}(\bar{s}))\\ &+\discount\sum_{s'}P(s'|\bar{s},\bar{a}_i,\policy_{-i}(\bar{s})){V}_i(s',\policy)\bigg)\bigg)\\
    =& \mu(s)\bigg(\stagePayoff_i(s,a_i,\policy_{-i}({s})) +\discount\sum_{s'}P(s'|{s},a_i,\policy_{-i}({s})){V}_i(s',\policy)\bigg)\\&+\delta \sum_{\bar{s}}\mu(\bar{s})\sum_{s'}P(s'|\bar{s},\policy)\frac{\partial {V}_i(s',\policy)}{\partial \policy_i(s,a_i)} \\ 
    =& \mu(s){Q}_i({s},a_i;\policy)+\delta \sum_{\bar{s}}\mu(\bar{s})\sum_{s'}P(s'|\bar{s},\policy)\frac{\partial {V}_i(s',\policy)}{\partial \policy_i(s,a_i)}\\
    =& \avg\ls{\mathbbm{1}(s^0=s)}{Q}_i({s},a_i;\policy) + \delta \avg\ls{ \frac{\partial {V}_i(s^1,\policy)}{\partial \policy_i(s,a_i)} }
\end{align*}}

We now suppose that the claim holds for some integer \(K\) and then show that it holds for \(K+1\), that is we have 
\begin{align*}
    &\frac{\partial {V}_i(\mu,\policy)}{\partial\policy_i(s,a_i)}=\avg\ls{\sum_{k=0}^{K}\discount^k \mathbbm{1}(s^k=s)} {Q}_i({s},a_i;\policy) \\&\hspace{2cm}+ \delta^{K+1}\avg\ls{ \frac{\partial {V}_i(s^{K+1},\policy)}{\partial \policy_i(s,a_i)}} \\
    &=\avg\ls{\sum_{k=0}^{K}\discount^k \mathbbm{1}(s^k=s)} {Q}_i({s},a_i;\policy)\\&+\delta^{K+1}\avg\ls{ \frac{\partial}{\partial \policy_i(s,a_i)}\lr{\sum_{a_i}\policy_i(s^{K+1},a_i){Q}_i(s^{K+1},a_i;\policy)} }
    \end{align*}
    \begin{align*}
&=\avg\ls{\sum_{k=0}^{K}\discount^k \mathbbm{1}(s^k=s)} {Q}_i({s},a_i;\policy)+\delta^{K+1}\Bigg( \mathbbm{1}(s^{K+1}=s) \\&\quad \cdot {Q}_i({s},a_i;\policy)+\delta\lr{\sum_{s'}P(s'|s^{K+1},\policy)\frac{\partial {V}_i(s',\policy)}{\partial \policy_i(s,a_i)}} \Bigg) \\ 
    &= \avg\ls{\sum_{k=0}^{K+1}\discount^k \mathbbm{1}(s^k=s)} {Q}_i({s},a_i;\policy) \\&\quad + \delta^{K+2}\avg\ls{ \frac{\partial {V}_i(s^{K+2},\policy)}{\partial \policy_i(s,a_i)}}.
\end{align*}
This completes the proof of \eqref{eq: InductionPG}. Now if we let \(K\ra\infty\) in \eqref{eq: InductionPG} then we obtain
\begin{align*}
    &\frac{\partial {V}_i(\mu,\policy)}{\partial\policy_i(s,a_i)}=\avg\ls{\sum_{k=0}^{\infty}\discount^k \mathbbm{1}(s^k=s)}Q_i(s,a_i;\pi) \\ &= \sum_{s^0\in S}\mu(s^0)\sum_{k=0}^{\infty}\Pr(s^k=s|s^0)Q_i(s,a_i;\pi) \\ 
    &= \frac{1}{1-\discount}d^{\policy}_{\mu}(s)Q_i(s,a_i;\pi).
\end{align*}
\noindent(b)
For any initial state distribution \(\mu\) and joint policy \(\policy=(\policy_i,\policy_{-i}), \policy'=(\policy_i',\policy_{-i})\in \Pi\), 
\begin{align*}
    &{V}_i(\mu,\policy)-{V}_i(\mu,\policy') \\&= \avg\ls{ \sum_{k=0}^{\infty}\discount^k \stagePayoff_i(s^k,a^k) } -{V}_i(\mu,\policy') \\ 
    =&\avg\bigg[\sum_{k=0}^{\infty}\discount^k \bigg( \stagePayoff_i(s^k,a^k)- {V}_i(s^k,\policy')+{V}_i(s^k,\policy') \bigg) \bigg] -{V}_i(\mu,\policy')
    \\ 
    =&\avg\ls{ \sum_{k=0}^{\infty}\discount^k \lr{ \stagePayoff_i(s^k,a^k) - {V}_i(s^k,\policy') } }   \\&+\avg\ls{{V}_i(s^0,\policy')}+ \avg\ls{\sum_{k=1}^{\infty} \delta ^{k}{V}_i(s^k,\policy') }-{V}_i(\mu,\policy'),
    \end{align*}
    where \(s^0\sim\mu,a^{k-1}\sim \policy(s^{k-1}), s^{k}\sim P(\cdot|s^{k-1},a^{k-1})\).
    We note that 
    \begin{align*}
&\avg\ls{{V}_i(s^0,\policy')} = {V}_i(\mu,\policy'), \\ &\avg\ls{\sum_{k=1}^{\infty} \delta ^{k}{V}_i(s^k,\policy')} = \delta\avg\ls{\sum_{k=0}^{\infty} \delta ^{k}{V}_i(s^{k+1},\policy') }.
    \end{align*}
    Therefore, 
  \begin{align*}
    &{V}_i(\mu,\policy)-{V}_i(\mu,\policy')\\&=\avg\ls{ \sum_{k=0}^{\infty}\discount^k \lr{ \stagePayoff_i(s^k,a^k)- {V}_i(s^k,\policy') } }  \\&\quad+\delta\avg\ls{\sum_{k=0}^{\infty} \delta ^{k}{V}_i(s^{k+1},\policy') }
    \\ 
    &=\avg\bigg[ \sum_{k=0}^{\infty}\discount^k \bigg( \stagePayoff_i(s^k,a^k) - {V}_i(s^k,\policy') +\discount {V}_i(s^{k+1},\policy') \bigg) \bigg]
    \\
&=\avg\bigg[\sum_{k=0}^{\infty}\discount^k \bigg( \stagePayoff_i(s^k,a^k)+\discount {V}_i(s^{k+1},\policy')
- {V}_i(s^k,\policy') \bigg) \bigg] \\
&=\avg\bigg[\sum_{k=0}^{\infty}\discount^k \bigg( \stagePayoff_i(s^k,a^k)+\discount \sum_{s'}P(s'|s^k,a^k){V}_i(s',\policy')\\
    &\quad- {V}_i(s^k,\policy') \bigg) \bigg].
\end{align*}
Thus, we conclude that 
\begin{align*}
     &{V}_i(\mu,\policy)-{V}_i(\mu,\policy')\\ 
    &= \avg\bigg[ \sum_{k=0}^{\infty}\discount^k \bigg( {Q}_i(s^k,a^k_i;\policy')- {V}_i(s^k,\policy') \bigg) \bigg]   \\ 
    &= \avg\ls{ \sum_{k=0}^{\infty}\discount^k  \advFunc_i(s^k,a^k_i;\policy')  } \\
    &= \frac{1}{1-\discount}\sum_{s'}d^{\pi}_{\mu}(s') \advFunc_i(s',\policy_i;\policy').
\end{align*}

{
\noindent (c)
    For every \(s\in S,\) define \(\Delta V_i(s,\pi) := V_i(s,\pi_{i},\pi_{-i}^{(\theta)})-V_i(s,\pi_{i},\pi_{-i})\). Since \(V_i(s,\pi_i,\pi_{-i}) = u_i(s,\pi_i,\pi_{-i}) + \discount \sum_{s'\in S}P(s'|s,\pi_i,\pi_{-i})V_i(s',\pi_i,\pi_{-i})\), we note that 
   {\begin{align}
        &\Delta V_i(s,\pi) = u_i(s,\pi_i,\pi_{-i}^{(\theta)})-u_i(s,\pi_i,\pi_{-i}) \notag \\ 
        & +\discount \sum_{s'\in S}P(s'|s,\pi_{i},\pi_{-i}^{(\theta)})V_i(s',\pi_{i},\pi_{-i}^{(\theta)})\notag \\ & -\discount \sum_{s'\in S}P(s'|s,\pi_{i},\pi_{-i})V_i(s',\pi_{i},\pi_{-i}) \notag \\ 
         &= u_i(s,\pi_i,\pi_{-i}^{(\theta)})-u_i(s,\pi_i,\pi_{-i})  \notag \\ 
        & + \discount \sum_{s'\in S}\lr{P(s'|s,\pi_i,\pi_{-i}^{(\theta)})-P(s'|s,\pi_{i},\pi_{-i})}V_i(s',\pi_{i},\pi_{-i}^{(\theta)}) \notag \\ 
        & + \discount \sum_{s'\in S}P(s'|s,\pi_{i},\pi_{-i})\lr{V_i(s',\pi_{i},\pi_{-i}^{(\theta)})-V_i(s',\pi_{i},\pi_{-i})},\notag 
    \end{align}}where the last equality is obtained by adding and subtracting the term \(\delta\sum_{s'}P(s'|s,\pi)V_i(s',\pi_i,\pi_{-i}^{(\theta)}).\)
Next, we note that \begin{align}\label{eq: Delta_V_i}
        &|\Delta V_i(s,\pi)| \leq |u_i(s,\pi_i,\pi_{-i}^{(\theta)})-u_i(s,\pi_i,\pi_{-i})| \notag \\ 
        & + \discount \bigg|\sum_{s'\in S}\lr{P(s'|s,\pi_i,\pi_{-i}^{(\theta)})-P(s'|s,\pi_{i},\pi_{-i})}V_i(s',\pi_{i},\pi_{-i}^{(\theta)})\bigg| \notag \\ 
        &+ \discount |\Delta V_i(s',\pi)|. 
    \end{align}

    First, for every \(s\in S, \pi\in \Pi\), we bound the term $|u_i(s,\pi_i,\pi_{-i}^{(\theta)})-u_i(s,\pi_i,\pi_{-i})|$ in \eqref{eq: Delta_V_i}. To bound this, we define a notation, for every \(i\in I,\)
    \begin{align*}
        \pi^{(\theta_{[1:k]})}_{-i} := \begin{cases}
            \pi_j^{(\theta)} & \text{if} \ j\in \{1,2,...k\}\backslash\{i\}, \\ 
            \pi_j  & \text{otherwise}.
        \end{cases}
    \end{align*}
    If \(k=0\) (resp. \(k=|\playerSet|\)) then \(\pi_{-i}^{(\theta_{[1:k]})}\) is \(\pi_{-i}\) (resp. \(\pi_{-i}^{(\theta)}\)). Using this notation, we obtain 
\begin{align}\label{eq: Equation_u_diff_Delta_V}
        &|u_i(s,\pi_i,\pi_{-i}^{(\theta)})-u_i(s,\pi_i,\pi_{-i})| \notag \\&= |\sum_{a_{-i}\in A_{-i}} u_i(s,\pi_i,a_{-i})(\textsf{Pr}^{\pi^{(\theta)}_{-i}}(a_{-i}|s) - \textsf{Pr}^{\pi_{-i}}(a_{-i}|s)) | \notag \\
        &= |\sum_{a_{-i}\in A_{-i}} u_i(s,\pi_i,a_{-i}) \notag \\ 
&\quad  \cdot\left(\sum_{k\in\playerSet\backslash\{i\}} (\textsf{Pr}^{\pi^{(\theta_{[1:k]})}_{-i}}(a_{-i}|s) - \textsf{Pr}^{\pi^{(\theta_{[1:k-1]})}_{-i}}(a_{-i}|s))\right) |\notag  \\ 
        &\leq  u_{\max}\sum_{a_{-i}\in A_{-i}} \sum_{k\in\playerSet\backslash\{i\}} |((1-\theta_k)\pi_k(a_{k}|s)\textsf{Pr}^{\pi^{(\theta_{[1:k-1]})}_{-ik}}(a_{-ik}|s)  \notag \\ & + \theta_k\frac{1}{|A_k|}\textsf{Pr}^{\pi^{(\theta_{[1:k-1]})}_{-ik}}(a_{-ik}|s) - \pi_k(a_k|s)\textsf{Pr}^{\pi^{(\theta_{[1:k-1]})}_{-ik}}(a_{-ik}|s)) | \notag 
        \\ 
        &=  u_{\max}\sum_{a_{-i}\in A_{-i}} \sum_{k\in\playerSet\backslash\{i\}} |(-\theta_k\pi_k(a_{k}|s)\textsf{Pr}^{\pi^{(\theta_{[1:k-1]})}_{-ik}}(a_{-ik}|s)  \notag \\ & + \theta_k\frac{1}{|A_k|}\textsf{Pr}^{\pi^{(\theta_{[1:k-1]})}_{-ik}}(a_{-ik}|s)) | \notag 
        \\ 
        &\leq u_{\max} \sum_{k\in\playerSet\backslash\{i\}}\theta_k \sum_{a_{-i}\in A_{-i}}\textsf{Pr}^{\pi^{(\theta_{[1:k-1]})}_{-i}}(a_{-i}|s) \notag   \\ &\quad \quad + u_{\max}\sum_{k\in\playerSet\backslash\{i\}} \theta_k\sum_{a_{-i}\in A_{-i}}\frac{1}{|A_k|} \textsf{Pr}^{\pi^{(\theta_{[1:k-1]})}_{-ik}}(a_{-ik}|s)  \notag
        \\
        &= 2u_{\max}\sum_{k\in\playerSet\backslash\{i\}}\theta_k,
    \end{align}
    where the last equality is using the fact that \(\textsf{Pr}^{\pi^{(\theta_{[1:k-1]})}_{-i}}(a_{-i}|s)\) and \(\textsf{Pr}^{\pi^{(\theta_{[1:k-1]})}_{-ik}}(a_{-ik}|s)\) are probability distribution on \(A_{-i}\) and \(A_{-ik}\) respectively. 

    Next, we bound the second term in \eqref{eq: Delta_V_i}. Note that 
    {\small\begin{align}
        &\bigg|\sum_{s'\in S}\lr{P(s'|s,\pi_i,\pi_{-i}^{(\theta)})-P(s'|s,\pi_{i},\pi_{-i})}V_i(s',\pi_{i},\pi_{-i}^{(\theta)})\bigg| \notag  \\ 
        &\leq \sum_{s'\in S}\bigg|\lr{P(s'|s,\pi_i,\pi_{-i}^{(\theta)})-P(s'|s,\pi_{i},\pi_{-i})}\bigg|\bar{V}_i \notag  \\ 
        &\leq \sum_{s'\in S}\bigg|\sum_{k\in \playerSet\backslash\{i\}}\Big(P(s'|s,\pi_i,\pi_{-i}^{(\theta_{[1:k]})})-P(s'|s,\pi_{i},\pi_{-i}^{(\theta_{[1:k-1]})})\Big)\bigg|\bar{V}_i\notag \\ 
        &= \bar{V}_i\sum_{s'\in S}\bigg|\sum_{k\in \playerSet\backslash\{i\}}\Big((1-\theta_k)P(s'|s,\pi_i,\pi_k,\pi_{-ik}^{(\theta_{[1:k-1]})})\notag \\ &\quad + \theta_k P(s'|s,\pi_i,\pi_k^{\circ},\pi_{-ik}^{(\theta_{[1:k-1]})}) -P(s'|s,\pi_{i},\pi_k,\pi_{-ik}^{(\theta_{[1:k-1]})})\Big)\bigg|\notag \\
        &= \bar{V}_i\sum_{s'\in S}\bigg|\sum_{k\in \playerSet\backslash\{i\}}(-\theta_k P(s'|s,\pi_i,\pi_k,\pi_{-ik}^{(\theta_{[1:k-1]})})\notag \\ &\quad + \theta_k P(s'|s,\pi_i,\pi_k^{\circ},\pi_{-ik}^{(\theta_{[1:k-1]})}))\bigg| \notag \\
        &\leq 2 \bar{V}_i\sum_{k\in \playerSet\backslash\{i\}}\theta_k,\label{eq: DElta_P_DeltaV}
    \end{align}}where \(\bar{V}_i = \max_{s'}|V_i(s',\pi_{i},\pi_{-i}^{(\theta)})|\) and the last inequality is using the fact that \(P(s'|s,\pi_i,\pi_k,\pi_{-ik}^{(\theta_{[1:k-1]})})\) and \(P(s'|s,\pi_i,\pi_k^{\circ},\pi_{-ik}^{(\theta_{[1:k-1]})})\) are probability distributions on \(S\).

    Combining \eqref{eq: Delta_V_i}, \eqref{eq: Equation_u_diff_Delta_V}, and \eqref{eq: DElta_P_DeltaV}, we obtain 
    \begin{align*}
        \max_{s,\pi}|\Delta V_i(s,\pi)| 
        &\leq \frac{\sum_{k\in \playerSet\backslash\{i\}}\theta_k}{1-\discount}\Big(2u_{\max}  + \frac{2\discount u_{\max}}{(1-\discount)}\Big) \\ &\leq\frac{2\sum_{k\in \playerSet\backslash\{i\}}\theta_k}{(1-\discount)^2}u_{\max}.
    \end{align*}
    This concludes the proof. 

\noindent (d)
    For any \(s\in S, i\in I,  a_i\in A_i, \pi\in \Pi\), we note that   
    {\begin{align*}
        &|Q_i(s,a_i;\pi_i,\pi_{-i}) -Q_i(s,a_i;\pi_i,\pi_{-i}^{(\theta)})| \\ 
        &\leq |u_i(s,a_i,\pi_{-i}) -u_i(s,a_i,\pi_{-i}^{(\theta)})| \\
        &\quad + \discount \bigg|\sum_{s'}P(s'|s,a_i,\pi_{-i})V_i(s'; \pi) \\ &\quad \quad - \sum_{s'}P(s'|s,a_i,\pi_{-i}^{(\theta)})V_i(s'; \pi_i,\pi_{-i}^{(\theta)})\bigg| \\ 
        &\stackrel{\eqref{eq: Equation_u_diff_Delta_V}}{\leq} 2\sum_{k\in \playerSet\backslash\{i\}}\theta_k u_{\max} + \discount \bigg|\sum_{s'}P(s'|s,a_i,\pi_{-i})V_i(s'; \pi) \\ &\quad \quad - \sum_{s'}P(s'|s,a_i,\pi_{-i}^{(\theta)})V_i(s'; \pi_i,\pi_{-i}^{(\theta)})\bigg| \\ 
        &\leq 2\sum_{k\in \playerSet\backslash\{i\}}\theta_k u_{\max}+ \discount \bigg|\sum_{s'}P(s'|s,a_i,\pi_{-i})\Big(V_i(s'; \pi)\\&\quad \quad  - V_i(s'; \pi_i,\pi_{-i}^{(\theta)})\Big)\bigg| + \discount \bigg| \sum_{s'} \Big(P(s'|s,a_i,\pi_{-i})\\
        &\quad-P(s'|s,a_i,\pi_{-i}^{(\theta)})\Big)V_i(s'; \pi_i,\pi_{-i}^{(\theta)})\bigg| \\ 
        &\stackrel{(i)}{\leq} 2\sum_{k\in \playerSet\backslash\{i\}}\theta_k u_{\max} + \discount \frac{2\sum_{k\in \playerSet\backslash\{i\}}\theta_k}{(1-\discount)^2}u_{\max} \\
        &\quad+ \delta \bigg| \sum_{s'} \Big(P(s'|s,a_i,\pi_{-i}) -P(s'|s,a_i,\pi_{-i}^{(\theta)})\Big)V_i(s'; \pi_i,\pi_{-i}^{(\theta)})\bigg| \\ 
        &\stackrel{\eqref{eq: DElta_P_DeltaV}}{\leq} u_{\max} \sum_{k\in \playerSet\backslash\{i\}}\theta_k \lr{ 2 + \discount \frac{2}{(1-\discount)^2} + \discount \frac{2}{(1-\discount)}} \\ 
        &= \frac{2\sum_{k\in \playerSet\backslash\{i\}}\theta_k }{(1-\discount)^2}u_{\max},
    \end{align*}}
    where \((i)\) is due to Lemma \ref{lem: TechnicalMain}-(c). This completes the proof. 
}

\subsection{Proof of Corollary \ref{cor: RefineConvergenceNew}}\label{ssec: Corollaryproof}
{
First, we note that for every \(\tilde{\epsilon} > 0\), there exists \(\epsilon > 0\) such that \(\epsilon + h_{\epsilon} = \tilde{\epsilon}\). This claim follows by noting that Assumption \ref{assm: SublevelSetNew} guarantees that the map \( \epsilon \in \mathbb{R}_+ \mapsto \epsilon + h_{\epsilon} \in \mathbb{R}_+\) is continuous function that goes to zero as \(\epsilon\) approaches \(0\). Furthermore, since \(h_{\epsilon}\) is non-decreasing in \(\epsilon\), we note that \(\epsilon + h_{\epsilon}\) is increasing in \(\epsilon
\). Next,  we show that every \(\tilde{\epsilon}, \tilde{\epsilon}'\) such that \(0 < \tilde{\epsilon} <   \tilde{\epsilon}',\) there exist positive scalars \(0<\epsilon< \epsilon'\) such that \({\epsilon}+h_{{\epsilon}} = \tilde{\epsilon}\) and \({\epsilon'}+h_{{\epsilon'}} = \tilde{\epsilon}'\). We show this by contradiction. Suppose for some \(0<\tilde{\epsilon} < \tilde{\epsilon}'\) it holds that \(\epsilon+h_{\epsilon}=\tilde{\epsilon}\) and \(\epsilon'+h_{\epsilon'}=\tilde{\epsilon'}\) with \(\epsilon \geq \epsilon'\). This implies that \(\tilde{\epsilon} = \epsilon + h_{\epsilon} \geq \epsilon' + h_{\epsilon'} = \tilde{\epsilon}'\), which contradicts the fact that \(\tilde{\epsilon} < \tilde{\epsilon}'\). 

Next, we show that for every \(\tilde{\epsilon} > 0\), the sequence of policies $\{\pi^t\}_{t=0}^{\infty}$ induced by Algorithm \ref{alg:independent_decentralized} converges to the set \(\textsf{NE}(\tilde{\epsilon})\) 
 with probability 1, if \(\sum_{i\in I}\theta_i < L\epsilon\), where \(\epsilon\) is such that \(\epsilon+h_{\epsilon} = \tilde{\epsilon}\). 
Following the same steps from the proof of Theorem \ref{theorem:independent}, it is sufficient to characterize the convergent set of the dynamical system \eqref{eq: policy_diff} in order to study the asymptotic behavior of the policy updates in Algorithm \ref{alg:independent_decentralized}. 

    From the proof of Lemma \ref{lemma: Policy_slow_continuous}, we know that  any absolutely continuous trajectory of \eqref{eq: policy_diff} converges to the set \(\Pi^\ast_\epsilon\), if \(\sum_{i\in I}\theta_i< L\epsilon\). 
    The proof concludes by noting that \(\Pi^\ast_{\epsilon} \subseteq \textsf{NE}(\epsilon+h_{\epsilon
    })=\textsf{NE}(\tilde\epsilon)\), due to Assumption \ref{assm: SublevelSetNew}.  
}

\section{Auxiliary Lemma}
{\begin{lemma}[\cite{leonardos2021global}]\label{lem: Grad_equal}
 Consider a Markov potential game \(\game,\) with potential function \(\Phi\). Then, for every \(s\in S, i\in I, a_i\in A_i, \pi\in\Pi\),
 \begin{align}\label{eq: Grad_equal}
     \frac{\partial V_i(s,\pi)}{\partial \pi_i(s,a_i)} = \frac{\partial \Phi(s,\pi)}{\partial \pi_i(s,a_i)}.
 \end{align}
\end{lemma}
\begin{proof}
    To prove this result, we first show that for every \(i\in I\), there exists a function \(U_i:S\times \Pi_{-i}\ra\mathbb{R}\) such that 
    \begin{align}\label{eq: To_Show}
        V_i(s,\pi) = \Phi(s,\pi) + U_i(s,\pi_{-i}), \quad \forall \ s\in S, \pi\in \Pi. 
    \end{align}
Fix arbitrary \(i\in I, \pi_i, \pi_i', \pi_i''\in \Pi_i\), and \(\pi_{-i}\in \Pi_{-i}\). By Definition \ref{def: PotentialGame}, it holds that, for every \(s\in S\),
    \begin{align*}
        \Phi(s,\pi_i,\pi_{-i}) - \Phi(s,\pi_i',\pi_{-i}) &= V_i(s,\pi_i,\pi_{-i}) - V_i(s,\pi_i',\pi_{-i})\\
        \Phi(s,\pi_i,\pi_{-i}) - \Phi(s,\pi_i'',\pi_{-i}) &= V_i(s,\pi_i,\pi_{-i}) - V_i(s,\pi_i'',\pi_{-i}).
    \end{align*}
    By re-arranging the terms in the above equation, we have
    \begin{equation}\label{eq: V_Phi_diff}
    \begin{aligned}
        &V_i(s,\pi_i,\pi_{-i}) - \Phi(s,\pi_i,\pi_{-i}) = V_i(s,\pi_i',\pi_{-i})- \Phi(s,\pi_i',\pi_{-i})\\
        &V_i(s,\pi_i,\pi_{-i}) - \Phi(s,\pi_i,\pi_{-i}) = V_i(s,\pi_i'',\pi_{-i})-\Phi(s,\pi_i'',\pi_{-i}).
    \end{aligned}
    \end{equation}
    Thus, equating the RHS in the above equation, we obtain that 
    \begin{align*}
        V_i(s,\pi_i',\pi_{-i})- \Phi(s,\pi_i',\pi_{-i}) = V_i(s,\pi_i'',\pi_{-i})-\Phi(s,\pi_i'',\pi_{-i}).  
    \end{align*}
    Since \(\pi_i'',\pi_i'\) are  arbitrary, we know that for every \(i\in I, s\in S, \pi_{-i}\in\Pi_{-i}\), \(
        V_i(s,\pi_i,\pi_{-i})-\Phi(s,\pi_i,\pi_{-i})
    \)
    does not depend on \(\pi_i\). Thus, using \eqref{eq: V_Phi_diff}, we conclude that \eqref{eq: To_Show} holds. 

    Equation \eqref{eq: Grad_equal} follows from taking the derivative with respect to \(\pi_i(s,a_i)\) on both sides of \eqref{eq: To_Show} and noting that \(U_i(s,\pi_{-i})\) does not depend on \(\pi_i\).
\end{proof}}

\begin{lemma}[Characterization of Nash equilibrium]  A policy \(\pi^\ast\in\Pi\) is a Nash equilibrium of \({\game}\) if and only if \(
    \pi^\ast(s) = {\mathrm{br}}_i(s;\pi^\ast)\) for all $i \in I$ and all $s \in S$. \label{lem:FixedPoint}
\end{lemma}
\begin{proof}
 We prove the claim in two parts -- first, we show that any policy \(\pi^\ast\) such that \(
    \pi^\ast(s) = {\mathrm{br}}_i(s;\pi^\ast)\) is a Nash equilibrium of \({\mathcal{G}}\). Next, we show the converse.

First, we provide an important characterization of the one-step optimal deviation which is crucial for the following proof
\begin{align}\label{eq: PolicyOpt}
    &{\mathrm{br}}_i(s;\pi_i^\ast, \pi_{-i}^{\ast,(\theta)})= \underset{\hat{\policy}_i\in \Delta(A_i)}{\arg\max}~\hat{\policy}_i^\top {Q}_i(s;\pi_i^\ast, \pi_{-i}^{\ast})\notag\\ 
&=\underset{\hat{\policy}\in \Delta(A_i)}{\arg\max}\bigg(\stagePayoff_i(s,\hat{\pi}_i,\pi_{-i}^{\ast})\\ &\quad \quad +\discount\sum_{s'}P(s'|s,\hat{\pi}_i,\pi_{-i}^{\ast}){V}_i(s',\pi_i^\ast, \pi_{-i}^{\ast})\bigg).
\end{align}

{First, we prove that $\tildeeq$ is a Nash equilibrium of $\game$, we need to show that for every \(i\in I, s\in S, \policy_i'\in \Pi_i\),
\begin{align}\label{eq:goal}
    {V}_i(s,\tildeeq_i,\pi_{-i}^{\ast}) \geq {V}_i(s,\policy_i',\pi_{-i}^{\ast}).
\end{align}

Before proving \eqref{eq:goal}, we first show that for any integer \(K\geq 1\), any \(s\in \stateSet\), any \(i\in\playerSet\), and any \(\policy_i'\in\Pi_i\),
\begin{align}\label{eq: ClaimVFunc}
    &{V}_i(s,\policy_i^\ast,\pi_{-i}^{\ast}) \notag \\ &\geq \avg \bigg[\sum_{k=0}^{K-1}\discount^k {u}_i(s^k,\policy_i',\pi_{-i}^{\ast}) + \delta^{K}{V}_i(s^{K},\tildeeq_i,\pi_{-i}^{\ast})\bigg],
\end{align}
where \(s^0=s, a_i^k\sim \policy_i'(s^k), a_{-i}^k \sim \policy^{*}_{-i}(s^k), s^k\sim P(\cdot|s^{k-1},a^{k-1})\). }

Consider \(K=1\), for any \(s\in \stateSet\), any \( i\in\playerSet\), any \(\policy'_i\in\Pi_i\) we have, 
\begin{align}\label{eq: ClaimBase}
&{\vFunc}_i(s,\tildeeq_i,\pi_{-i}^{\ast})\notag  \\
     &= {{\stagePayoff}_i(s,\policy_i^\ast,\pi_{-i}^{\ast}) +\discount\sum_{s'}P(s'|s,\tildeeq_i,\pi_{-i}^{\ast}){V}(s',\tildeeq_i, \pi_{-i}^{\ast})} 
\\
     &\geq {\stagePayoff}_i(s,\policy_i',\pi_{-i}^{\ast}) +\discount\sum_{s'}P(s'|s,\policy_i',\tildeeq_{-i}){V}(s',\tildeeq_i,\pi_{-i}^{\ast}) \notag \\
     &= \avg\ls{{u}_i(s^0,\policy_i',\pi_{-i}^{\ast,(\theta)}) + \delta {V}_i(s^1,\tildeeq_i,\pi_{-i}^{\ast,(\theta)}) },\notag
\end{align}
where, again, $s^0=s$, $a_i^0\sim \policy_i^{'}(s^0)$, $a_{-i}^0 \sim \policy^{*}_{-i}(s^0)$, $s^1\sim P(\cdot|s^{0},a^{0})$ and the inequality follows from \eqref{eq: PolicyOpt} as \(\pi_i^\ast(s) \in \br_i(s;\pi^\ast)\) for every \(i\in I, s\in S\).

Next, suppose that \eqref{eq: ClaimVFunc} holds for some integer \(K\),  we consider $K+1$: 
\begin{align*}
&{V}_i(s,\policy_i^\ast,\pi_{-i}^{\ast}) \notag \\ \underset{(a)}{\geq} &\avg\ls{ \sum_{k=0}^{K-1}\discount^k {u}_i(s^k,\policy_i',\pi_{-i}^{\ast}) + \delta^{K}{V}_i(s^{K},\tildeeq_i,\pi_{-i}^{\ast})} \\ 
    \underset{(b)}{=}& \avg\bigg[ \sum_{k=0}^{K-1}\discount^k {u}_i(s^k,\policy_i',\pi_{-i}^{\ast}) + \delta^{K}\bigg( {u}_i(s^K,\policy_i^\ast,\pi_{-i}^{\ast}) \\&\hspace{1cm}+ \delta \sum_{s'}P(s'|s^K,\policy_i^\ast,\pi_{-i}^{\ast}){V}_i(s',\tildeeq_i,\pi_{-i}^{\ast}) \bigg)\bigg] \\ 
    \underset{(c)}{\geq}& \avg\bigg[ \sum_{k=0}^{K-1}\discount^k {u}_i(s^k,\policy_i',\pi_{-i}^{\ast}) + \delta^{K}\bigg( {u}_i(s^K,\policy_i',\pi_{-i}^{\ast}) \\&\hspace{1cm}+ \delta \sum_{s'}P(s'|s^K,\policy_i',\pi_{-i}^{\ast}){V}_i(s',\tildeeq_i,\pi_{-i}^{\ast}) \bigg)\bigg] \\ 
    \underset{(d)}{=}& \avg\ls{ \sum_{k=0}^{K}\discount^k {u}_i(s^k,\policy_i',\pi_{-i}^{\ast}) + \delta^{K+1}{V}_i(s^{K+1},\tildeeq_i,\pi_{-i}^{\ast})},
\end{align*}
where \((a)\) is by induction hypothesis, \((b)\) is due to \eqref{eq: ClaimBase}, \((c)\) is due to \eqref{eq: PolicyOpt}
and \((d)\) is by rearrangement of terms. Thus, by mathematical induction, we have established that \eqref{eq: ClaimVFunc} holds for all $K$. Let \(K\ra\infty\) in \eqref{eq: ClaimVFunc}, we have  
\begin{align*}
{V}_i(s,\policy_i^\ast,\pi_{-i}^{\ast}) \geq \avg\ls{ \sum_{k=0}^{\infty}\discount^k {u}_i(s^k,\policy_i',\pi_{-i}^{\ast})} = {V}_i(s,\policy_i',\pi_{-i}^{\ast}),
\end{align*}
for every  \(s\in\stateSet, i\in\playerSet, \policy_i'\in\Pi\). 
Thus, we have proved \eqref{eq:goal}, i.e. \(\tildeeq\) is a Nash equilibrium of game \({\mathcal{G}}\).

Next, we show that any Nash equilibrium \({\pi}^\ast\) of \({\mathcal{G}}\) satisfies that \(\tildeeq_i(s) \in {\mathrm{br}}_i(s;\tildeeq)\) for every \(i\in I, s\in S\). We prove this by contradiction. Suppose there exists a player \(i\in I\), and set of states \(\bar{S}\subset S\) such that for every \(\bar{s}\in \bar{S}\) it holds that \(\tildeeq_i(\bar{s}) \not\in  {\mathrm{br}}_i(\bar{s};\tildeeq)\). Let \(\pi'\) be a policy such that for all \(s\in S, i\in I\), \(\pi_i'(s) \in {\mathrm{br}}_i(s;\tildeeq)\). 
Without loss of generality we assume \(|\bar{S}|=1\).

We claim that for any integer \(K\geq 1\), any \(s\in S\), \(i\in I\) it holds that 
\begin{align}\label{eq: ClaimVFuncCont}
&{V}_i(s,\policy_i^\ast,\pi_{-i}^{\ast}) \notag \\ & \leq \avg\ls{ \sum_{k=0}^{K-1}\discount^k {u}_i(s^k,\policy_i',\pi_{-i}^{\ast}) + \delta^{K}{V}_i(s^{K},\tildeeq_i,\pi_{-i}^{\ast})},
\end{align}
where \(s^0=s, a_i^k\sim \policy_i'(s^k), a_{-i}^k \sim \policy^{*}_{-i}(s^k), s^k\sim P(\cdot|s^{k-1},a^{k-1})\) and the inequality is strict for \(s=\bar{s}\).

Consider \(K=1\), for any \(s\in \stateSet\), any \( i\in\playerSet\), we have 
\begin{align*}
    &{V}_i(s,\policy_i^\ast,\pi_{-i}^{\ast}) \\ 
    &= {u}_i(s,\tildeeq_i,\pi_{-i}^{\ast}) +\delta \sum_{s'}P(s'|s,\tildeeq_i,\pi_{-i}^{\ast}){V}_i(s',\tildeeq_i,\pi_{-i}^{\ast})\\
    &\leq {u}_i(s,\pi_i',\tildeeq_{-i}) +\delta \sum_{s'}P(s'|s,\pi_i',\tildeeq_{-i}){V}_i(s',\tildeeq) \\
    &= \avg\ls{{u}_i(s^0,\policy_i',\policy_{-i}^\ast) + \delta {V}_i(s^1,\tildeeq) },\notag
\end{align*}
where, again, $s^0=s$, $a_i^0\sim \policy_i'(s^0)$, $a_{-i}^0 \sim \policy^{*}_{-i}(s^0)$, $s^1\sim P(\cdot|s^{0},a^{0})$ and the inequality follows from \eqref{eq: PolicyOpt}. Note that inequality is strict for \(s^0=\bar{s}\).

Next, suppose \eqref{eq: ClaimVFuncCont} holds for some integer \(K\), we consider \(K+1\):
\begin{align*}
&{V}_i(s,\policy_i^\ast,\policy_{-i}^\ast) \underset{(a)}{\leq} \avg\ls{ \sum_{k=0}^{K-1}\discount^k {u}_i(s^k,\policy_i',\policy_{-i}^\ast) + \delta^{K}{V}_i(s^{K},\tildeeq)} \\ 
{=}& \avg\bigg[ \sum_{k=0}^{K-1}\discount^k {u}_i(s^k,\policy_i',\policy_{-i}^\ast) + \delta^{K}\bigg( {u}_i(s^K,\policy_i^\ast,\policy_{-i}^\ast) \\&\hspace{2cm}+ \delta \sum_{s'}P(s'|s^K,\policy_i^\ast,\policy_{-i}^\ast){V}_i(s',\tildeeq) \bigg)\bigg] 
 \\ 
    \underset{(b)}{\leq}& \avg\bigg[ \sum_{k=0}^{K-1}\discount^k {u}_i(s^k,\policy_i',\policy_{-i}^\ast) + \delta^{K}\bigg( {u}_i(s^K,\policy_i',\policy_{-i}^\ast) \\&\hspace{2cm}+ \delta \sum_{s'}P(s'|s^K,\policy_i',\policy_{-i}^\ast){V}_i(s',\tildeeq) \bigg)\bigg] \\ 
    \underset{(c)}{=}& \avg\ls{ \sum_{k=0}^{K}\discount^k {u}_i(s^k,\policy_i',\policy_{-i}^\ast) + \delta^{K+1}{V}_i(s^{K+1},\tildeeq)},
\end{align*}
where \((a)\) is by induction hypothesis, \((b)\) is due to \eqref{eq: PolicyOpt}
and \((c)\) is by rearrangement of terms. Thus, by mathematical induction, we have established that \eqref{eq: ClaimVFuncCont} holds for all $K$. Let \(K\ra\infty\) in \eqref{eq: ClaimVFuncCont}, we have  
\begin{align*}
{V}_i(s,\policy_i^\ast,\policy_{-i}^\ast) \leq \avg\ls{ \sum_{k=0}^{\infty}\discount^k {u}_i(s^k,\policy_i',\policy_{-i}^\ast)} = {V}_i(s,\policy_i',\policy_{-i}^\ast),
\end{align*}
for every  \(s\in\stateSet, i\in\playerSet, \policy_i'\in\Pi\). Furthermore, 
\begin{align*}
{V}_i(\bar{s},\policy_i^\ast,\policy_{-i}^\ast) < \avg\ls{ \sum_{k=0}^{\infty}\discount^k {u}_i(s^k,\policy_i',\policy_{-i}^\ast)} = {V}_i(\bar{s},\policy_i',\policy_{-i}^\ast),
\end{align*}

This contradicts the fact that \({\pi}_i^\ast\) is a Nash equilibrium of game \(\game\).
\end{proof}

\end{document}